\documentclass{article}

\usepackage{microtype}
\usepackage{graphicx}
\usepackage{subfigure}
\usepackage{booktabs} 

\usepackage{hyperref}

\usepackage[accepted]{icml2023}

\usepackage{amsmath}
\usepackage{amssymb}
\usepackage{mathtools}
\usepackage{amsthm}

\usepackage[capitalize,noabbrev]{cleveref}

\theoremstyle{plain}

\theoremstyle{definition}

\theoremstyle{remark}

\usepackage[textsize=tiny]{todonotes}
\usetikzlibrary{calc, shapes.multipart, backgrounds}

\newcommand{\fop}[2]{F_{#1}^{#2}}
\newcommand{\bop}[2]{B_{#1}^{#2}}
\newcommand{\act}[1]{x_{#1}}
\newcommand{\bact}[2]{\bar{x}_{#1}^{#2}}
\newcommand{\mtmp}[1]{tmp(#1)}
\newcommand{\exec}[1]{r(#1)}
\newcommand{\opt}{\text{Opt}}

\icmltitlerunning{Rockmate: an Efficient, Fast, Automatic and Generic Tool for Re-materialization in PyTorch}

\begin{document}

\twocolumn[
    \icmltitle{Rockmate: an Efficient, Fast, Automatic and Generic Tool \\for Re-materialization in PyTorch}



    \icmlsetsymbol{equal}{*}

    \begin{icmlauthorlist}
        \icmlauthor{Xunyi Zhao}{equal,inria}
        \icmlauthor{Théotime Le Hellard}{equal,ens}
        \icmlauthor{Lionel Eyraud-Dubois}{inria}
        \icmlauthor{Julia Gusak}{inria}
        \icmlauthor{Olivier Beaumont}{inria}
    \end{icmlauthorlist}

    \icmlaffiliation{inria}{Inria Center at the University of Bordeaux}
    \icmlaffiliation{ens}{École Normale Supérieure, PSL University, Paris}


    \icmlcorrespondingauthor{Xunyi Zhao}{xunyi.zhao@inria.fr}
    \icmlcorrespondingauthor{Lionel Eyraud-Dubois}{lionel.eyraud-dubois@inria.fr}

    \icmlkeywords{Machine Learning, ICML}

    \vskip 0.3in
]



\printAffiliationsAndNotice{\icmlEqualContribution} 

\begin{abstract}
    We propose Rockmate to control the memory requirements when training
    PyTorch DNN models. Rockmate is an automatic tool that starts from
    the model code and generates an equivalent model, using a predefined
    amount of memory for activations, at the cost of a few re-computations.
    Rockmate automatically detects the structure of computational
    and data dependencies and rewrites the initial model as a sequence of
    complex blocks. We show that such a structure is widespread and can be
    found in many models in the literature (Transformer based models, ResNet,
    RegNets,...). This structure allows us to solve the problem in a fast
    and efficient way, using an adaptation of Checkmate (too slow on the
    whole model but general) at the level of individual blocks and an
    adaptation of Rotor (fast but limited to sequential models) at the level
    of the sequence itself. We show through experiments on many models
    that Rockmate is as fast as Rotor and as efficient as Checkmate,
    and that it allows in many cases to obtain a significantly
    lower memory consumption for activations (by a factor of 2 to 5)
    for a rather negligible overhead (of the order of 10\% to 20\%).
    Rockmate is open source and available at \url{https://github.com/topal-team/rockmate}.
\end{abstract}

\section{Introduction}
\label{sec:intro}

%
%
%
%
    In recent years, very large networks have emerged.
    These networks induce huge memory requirements both
    because of the number of parameters and
    the size of the activations that must be kept in memory to perform
    back-propagation.
    Memory issues for training have been identified for a long time. Indeed,
    training is usually performed on computing resources such as GPUs or TPUs, on
    which memory is limited. Therefore, different approaches have been proposed.

The first category of solutions consists in relying on {\bf parallelism}. Data
parallelism allows to distribute the memory related to the activations, at the
cost of exchanging the network weights between the different resources using
collective communications such as {\tt MPI\_AllReduce} which can be expensive
for networks such as those of the GPT2 class. On the contrary, model
parallelism allows to distribute the weights of the network, at the cost of the
communication of activations and memory overheads in case it is used in a
pipelined way, and its scalability is limited by nature.

The second category of solutions is purely sequential. {\bf Offloading} makes it
possible to move some activations computed during the forward phase from the
memory of the accelerator (GPU or TPU) to the memory of the CPU, and then to
prefetch them back at the appropriate moment into the memory of the GPU during
the backward phase. This solution therefore consumes bandwidth on the PCI-e bus
between the CPU and the accelerator, which is also used to load training data.
Another solution, called {\bf re-materialization}, consists in deleting from
accelerator memory some activations computed during the forward phase and then
recomputing them during the backward phase. This approach does not consume
communication resources, but it does induce a computational overhead.

In the present paper, we focus on the latter re-materialization approach on a
single GPU or TPU, which is sufficient in practice for the size of the networks
we consider in the experiments and which can be trivially combined with data parallelism to
accelerate training. In this framework, for a given memory constraint, the {\bf optimization problem consists in finding a
    sequence of computing, forgetting and
    recomputing actions which allows to perform the training for given
    inputs
    and batch sizes, while fulfilling the memory constraint and minimizing the
    computational overhead.}
%
%

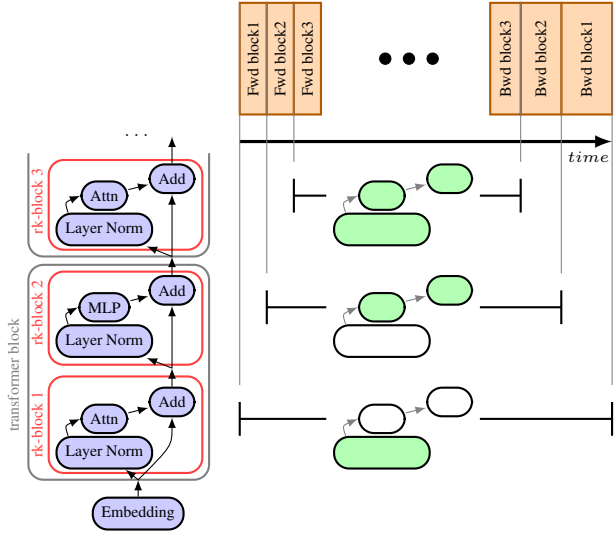
\begin{figure}
  \centering
  \definecolor{brass}{rgb}{0.71, 0.65, 0.26}
\begin{tikzpicture}[every node/.style={font=\tiny}, scale=0.45, >=latex]
  \tikzset{
    box/.style={thick, rounded corners=.2cm},
    operation/.style={box, fill=#1, draw=black},
    block/.style={box, draw=red!75},
    tblock/.style={box, draw=gray},
    fwd/.style={thick, draw=fwd!70!black, fill=fwd!30},
    bwd/.style={thick, draw=bwd!70!black, fill=bwd!30},
    saved/.style={fill=green!30},
    notsaved/.style={fill=white},
    dot/.style={anchor=center, inner sep=0cm, minimum width=.15cm, circle, fill=black, draw},
  }
  \colorlet{op}{blue!20}
  \colorlet{embed}{op}
  \colorlet{ln}{op}
  \colorlet{attn}{op}
  \colorlet{mlp}{op}
  \colorlet{add}{op}
  \colorlet{fwd}{orange}
  \colorlet{bwd}{orange}
  \begin{scope}
    \node[operation=embed] (embed) at (0, 0) {Embedding};

    \begin{scope}[yshift=0.3cm]
      \coordinate (input1) at (0, 0.7);
      \node[operation=ln] (ln1) at (-1, 1.5) {Layer Norm};
      \node[operation=attn] (attn1) at (-1, 2.5) {Attn};
      \node[operation=add] (add1) at (1, 3) {Add};
    \end{scope}

    \begin{scope}[yshift=3.6cm]
      \coordinate (input2) at (1, .7);
      \node[operation=ln] (ln2) at (-1, 1.5) {Layer Norm};
      \node[operation=mlp] (mlp) at (-1, 2.5) {MLP};
      \node[operation=add] (add2) at (1, 3) {Add};
    \end{scope}

    \begin{scope}[yshift=6.9cm]
      \coordinate (input3) at (1, .7);
    \node[operation=ln] (ln3) at (-1, 1.5) {Layer Norm};
    \node[operation=attn] (attn3) at (-1, 2.5) {Attn};
    \node[operation=add] (add3) at (1, 3) {Add};
    \end{scope}

    \begin{scope}[yshift=9.7cm]
      \node (cont) at (0, 1.5) {$\cdots$};
    \end{scope}

    \draw[block] ($(ln1.south west) + (-0.2, -0.1)$) rectangle ($(add1.north east) + (0.2, .3)$);
    \draw[block] ($(ln3.south west) + (-0.2, -0.1)$) rectangle ($(add3.north east) + (0.2, .1)$);
    \draw[block] ($(ln2.south west) + (-0.2, -0.2)$) rectangle ($(add2.north east) + (0.2, .1)$);

    \node[rotate=90,color=red!75, anchor=south west] at ($(ln1.west) + (-0.1, -0.4)$) {rk-block 1};
    \node[rotate=90,color=red!75, anchor=south west] at ($(ln2.west) + (-0.1, -0.4)$) {rk-block 2};
    \node[rotate=90,color=red!75, anchor=south west] at ($(ln3.west) + (-0.1, -0.4)$) {rk-block 3};
    \node[rotate=90,color=gray, anchor=south] at ($(ln2.south west) + (-0.8, -0.5)$) {transformer block};
    
    \draw[tblock] ($(ln1.south west) + (-0.8, -0.3)$) rectangle ($(add2.north east) + (0.4, .3)$);
    \draw[tblock] (perpendicular cs: vertical line through={($(ln3.south west) + (-0.8, -0.3)$)},
                                     horizontal line through={($(add3.north) + (0, 0.3)$)})
                                     -- ($(ln3.south west) + (-0.8, -0.3)$)
                                     -| ($(add3.north east) + (0.4, .3)$);

   \begin{scope}[->]
     \draw (embed) -- (input1);
      \draw (input1) -- (ln1);
      \draw (input1) ..controls +(1, 1).. (add1);
      \draw (ln1.155) ..controls +(-0.1, 0.3).. (attn1.west);
      \draw (attn1) -- (add1);

      \draw (add1) -- (input2);
      \draw[] (input2) -- (ln2);
      \draw (input2) -- (add2);
      \draw (ln2.155) ..controls +(-0.1, 0.3).. (mlp.west);
      \draw (mlp) -- (add2);

      \draw (add2) -- (input3);
      \draw[] (input3) -- (ln3);
      \draw (input3) -- (add3);
      \draw (ln3.155) ..controls +(-0.1, 0.3).. (attn3.west);
      \draw (attn3) -- (add3);

      \draw (add3) -- +(0, 1.25);
    \end{scope}
  \end{scope}

  \begin{scope}[xshift=3cm] 
    \begin{scope}[yshift=11cm]
      \draw[very thick, ->] (0, 0) -- +(11, 0) node[pos=1.02, anchor=north east]{$time$};
      \foreach \x in {1, 2, 3} {
         \draw[fwd] (\x*.8, .8) rectangle +(-.8, 3.3) node[midway, rotate=90] {Fwd block\x} coordinate (F\x);
      }
      \begin{scope}[xshift=3.6cm]
        \foreach \x in {1, 2, 3} {
          \node[dot] at (\x*.7, 2.45) {};
        }
     \end{scope}
     
     \begin{scope}[xshift=1cm]
       \draw[bwd] (8.5, 0.8) rectangle +(1.5, 3.3) node[midway, rotate=90] {Bwd block1} coordinate (B1);
        \draw[bwd] (7.3, .8)  rectangle +(1.2, 3.3) node[midway, rotate=90] {Bwd block2} coordinate (B2);
        \draw[bwd] (6.4, .8) rectangle +(.9, 3.3) node[midway, rotate=90] {Bwd block3} coordinate (B3);
     \end{scope}
    \end{scope}
    \begin{scope}[text opacity=0, xshift=5.2cm]
      \begin{scope}[yshift=0.3cm]
        \node[operation=ln, saved] (ln) at (-1, 1.5) {Layer Norm};
        \node[operation=attn, notsaved] (lay) at (-1, 2.5) {Attn};
        \node[operation=add, notsaved] (add) at (1, 3) {Add};
        \draw[->, gray] (ln.155) ..controls +(-0.1, 0.3).. (lay.west);
        \draw[->, gray] (lay) -- (add);
        \coordinate (s) at (F1 |- lay.center);
        \coordinate (e) at (B1 |- lay.center);
        \draw[gray] (F1) -- ($(s)!1cm!(F1)$) (B1) -- ($(e)!1cm!(B1)$);
        \draw[thick] (lay.center -| ln.west) ++(-.2, 0) -- (F1 |- lay.center)
         (lay.center -| add.east) ++(.2, 0) -- (B1 |- lay.center);
         \draw[thick] ($(s)!.5cm!(F1)$) -- ($(s)!-.5cm!(F1)$) ($(e)!.5cm!(B1)$) -- ($(e)!-.5cm!(B1)$);
      \end{scope}

      \begin{scope}[yshift=3.6cm]
        \node[operation=ln, notsaved] (ln) at (-1, 1.5) {Layer Norm};
        \node[operation=mlp, saved] (lay) at (-1, 2.5) {Attn};
        \node[operation=add, saved] (add) at (1, 3) {Add};
        \draw[->, gray] (ln.155) ..controls +(-0.1, 0.3).. (lay.west);
        \draw[->, gray] (lay) -- (add);
        \coordinate (s) at (F2 |- lay.center);
        \coordinate (e) at (B2 |- lay.center);
        \draw[gray] (F2) -- ($(s)!1cm!(F2)$) (B2) -- ($(e)!1cm!(B2)$);
        \draw[thick] (lay.center -| ln.west) ++(-.2, 0) -- (F2 |- lay.center)
         (lay.center -| add.east) ++(.2, 0) -- (B2 |- lay.center);
         \draw[thick] ($(s)!.5cm!(F2)$) -- ($(s)!-.5cm!(F2)$) ($(e)!.5cm!(B2)$) -- ($(e)!-.5cm!(B2)$);
      \end{scope}

      \begin{scope}[yshift=6.9cm]
        \node[operation=ln, saved] (ln) at (-1, 1.5) {Layer Norm};
        \node[operation=attn, saved] (lay) at (-1, 2.5) {Attn};
        \node[operation=add, saved] (add) at (1, 3) {Add};
        \draw[->, gray] (ln.155) ..controls +(-0.1, 0.3).. (lay.west);
        \draw[->, gray] (lay) -- (add);
        \coordinate (s) at (F3 |- lay.center);
        \coordinate (e) at (B3 |- lay.center);
        \draw[gray] (F3) -- ($(s)!1cm!(F3)$) (B3) -- ($(e)!1cm!(B3)$);
        \draw[thick] (lay.center -| ln.west) ++(-.2, 0) -- (F3 |- lay.center)
         (lay.center -| add.east) ++(.2, 0) -- (B3 |- lay.center);
         \draw[thick] ($(s)!.5cm!(F3)$) -- ($(s)!-.5cm!(F3)$) ($(e)!.5cm!(B3)$) -- ($(e)!-.5cm!(B3)$);
      \end{scope}
    \end{scope}
    
  \end{scope}
  
\end{tikzpicture}
  \caption{Simplified example of running Rockmate on a GPT
    model. \textbf{Left:} Dependency graph of the first part of the
    model, where transformer blocks are shown in gray, and Rockmate
    blocks are identified in red. \textbf{Right:} (top) a schedule
    corresponding to the first three blocks; (bottom) indication of
    which activations are saved (green) or not (white) for each block,
    and the intervals during which they are present in memory. Saving
    fewer activations leads to more recomputation and thus longer
    backward time.}
  \label{fig:example.gpt}
\end{figure}

To find the optimal sequence, different approaches have been
proposed. In the first approach, like in
Rotor~\cite{rotor-RR}, it is assumed that the dependencies within the model
have a particular structure, typically a sequence of
operations. In this case, using dynamic programming, it is possible to
find the optimal order of computations in reasonable time. On the other hand, in
the case where the computations performed by the model do not
naturally consist in a sequence of operations, this approach requires
to aggregate elementary operations into complex blocks to make the
chain structure emerge. In this case, re-materialization decisions
have to be made at the level of blocks, which reduces optimization
opportunities. The left of Figure~\ref{fig:example.gpt} shows the graph of a GPT-like model, where each block
corresponds to one half of a transformer block. On such a graph, this
approach has to decide during the forward phase whether to keep all 
internal activations or to delete all of them (and to recompute them during
the backward phase).


\begin{figure}
    \centering
    \includegraphics[width=\linewidth]{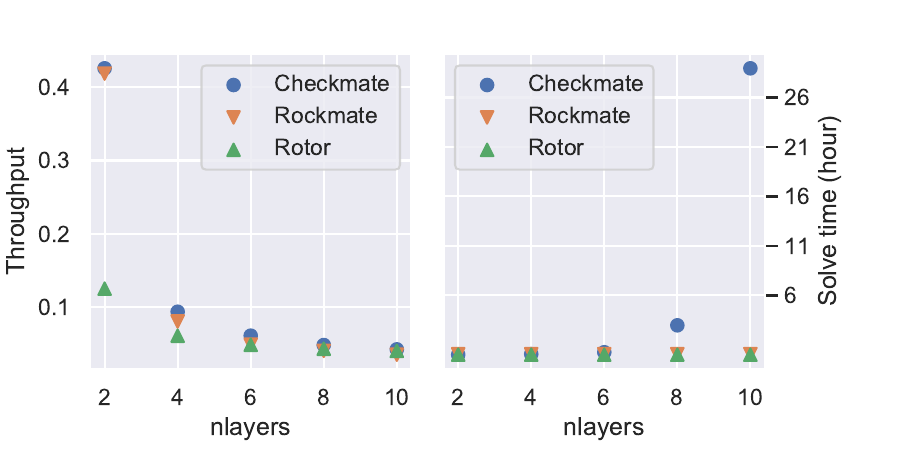}
    \caption{Efficiency comparison between Rockmate, Rotor and
        Checkmate for GPT2 networks with $nlayers$ transformer blocks. 
        Rockmate finds an optimized re-materialization solution as quickly as Rotor while keeping
        a similar performance as Checkmate in terms of resulting throughput.
        Throughput is defined as the number of samples processed per time unit (ms).
    }
    \label{fig:checkmate_lin}
\end{figure}

In the case of general graphs that are not structured as a sequence of
elementary operations, another approach has been proposed in
Checkmate~\cite{jain2020checkmate}. It consists in describing the
operations corresponding to both forward and backward phases as a
Directed Acyclic Graph (DAG) and to find the optimal solution through
solving an Integer Linear Program (ILP).
The number of integer
variables is proportional to $V \times E$, where $V$ is the number of
operations and $E$ is the number of arcs of the DAG. Hence, a major
shortcoming of this approach is the computational time induced by
solving the ILP. Typically, even using commercial solvers such as
CPLEX or Gurobi, it is not possible (in one day of computation) to
consider GPT2 models with more than 10 transformer blocks (see Figure~\ref{fig:checkmate_lin}),
while classical instances include several dozens.


In the present paper {\bf we propose Rockmate, a new re-materialization strategy,}
in which models are seen as a sequence of blocks (in the sense of
Rotor), but where several optimal strategies are pre-computed for each
block (using a Checkmate-like approach). A simple example of the
resulting execution is shown on the right of
Figure~\ref{fig:example.gpt}, where in each block, a different set of
activations is saved, resulting in different backward execution
times. In reality, for a GPT model, Rockmate divides each block into 9
or 6 operations for the first or second half of the transformer block
respectively, and the execution can also contain re-executions of some
blocks.

As our experimental results demonstrate, {\bf for a
large variety of networks Rockmate can compute
near-optimal solutions (close to Checkmate quality in terms of
throughput) in a reasonable time (close to Rotor runtime, faster than Checkmate), by
combining the advantages of both approaches}. A preview shown on
Figure~\ref{fig:checkmate_lin} presents the throughput and
solving time of all three solutions for GPT neural networks with
varying number of transformer blocks.


Another contribution is that {\bf we have built a framework}, 
which can be easily applied on any PyTorch \texttt{nn.Module}. It contains
complete implementation of our
algorithm (Algorithm~\ref{alg:Rockmate}), including main phases with {\bf newly proposed
computation-data graph builder, integer linear programming, and dynamic programming 
techniques} described in section~\ref{sec:models}.
Rockmate takes the model as input and automatically builds the data-flow
graph with measurements (computation time, output size and peak memory
of each operation, etc.). The optimal schedule is then determined based on the graph
and used to build a new \texttt{nn.Module} which runs forward and backward phases
within a given memory constraint. In Section~\ref{sec:experiments}, we demonstrate that 
the resulting new GPT2 models can achieve the same result as the original ones
with 25\% computational overhead, while using only 25\% of the original memory
needs to store the activations.




Note that {\bf all the benefits
of Rockmate do not induce any accuracy loss for the model}:
given the same batch of training data, the Rockmate model will
compute exactly the same gradient values for every trainable parameter compared to the
original model. Hence, both models achieve the same accuracy after the same number
of training epochs.



\section{Related works}
\label{related_works}



During training, memory requirements are very demanding. On the one hand, they
come from the storage of the network weights, and the associated intermediate data,
such as gradients and optimizer states. On the other hand, memory requirements also come from the
storage of the activations associated with gradient descent, since (almost) all
the results computed during the forward phase must be kept in memory until they
are used by the gradient computation during the backward phase.

There are different strategies for saving memory when training Deep Neural Networks (DNNs), adapted to
these different memory requirements. We can differentiate between strategies
that rely on the use of parallelism (data parallelism, model parallelism),
those that use the possibility of transferring data to another device than the
memory of the accelerator (denoted as offloading or paging in the literature),
and those that rely on the redundant computation of activations deleted from
the memory (denoted as checkpointing or re-materialization in the literature).
Of course, these strategies can naturally be combined since they rely on
different resource consumption (use of several computing resources for
parallelism, external storage for offloading or re-computations for
re-materialization). This combination is more or less difficult because some
resources are consumed by several approaches (computations on the GPU or the
TPU of course, but also communications on the PCI-e bus or on the NVLink). In
the rest of this section, we will focus mainly on re-materialization
strategies after having briefly discussed the other approaches.

Among the most popular parallel strategies for DNN training are data
parallelism and model parallelism. Data parallelism is based on the idea of
performing forward and backward phases in parallel on different data and on
several GPUs. The gradients computed on the different GPUs must then be
reduced, which requires collective communication of all the weights. This
approach used in isolation was very popular in more or less
synchronous~\cite{zinkevich2010parallelized,datap} or
asynchronous~\cite{zhang2013asynchronous} variants for convolutional networks
in which the network weights were small compared to the activations, but is nowadays
mainly used in combination with model parallelism. Model parallelism
consists in distributing layers and weights over different resources and
communicating forward activations and their gradients between GPUs. This
approach has been popularized in frameworks like GPipe~\cite{huang2019gpipe},
Pipedream~\cite{pipedream} or Varuna~\cite{varuna} and its complexity has been
studied in~\cite{beaumont2021pipelined}. It is often combined with systematic
re-materialization such as in GPipe~\cite{huang2019gpipe}.

Offloading (sometimes denoted as paging) can be applied to both network weights
and activations. The idea is simply to remove the memory load from the GPU and
store data in CPU memory, that is typically much larger than GPU memory. This
idea is relevant both for
activations, computed during the
forward phase but which will not be used for a long time by the backward phase~\cite{vdnn1,le2018tflms,beaumont2020gpucpu}
and for network weights~\cite{beaumont:hal-03580767}, which are also used only
once during the forward phase and once during the backward. Offloading can also
be combined with re-materialization~\cite{beaumont2021efficient}, which is
particularly relevant for decentralized training on tiny
devices~\cite{patil2022poet}.

Historically, re-materialization strategies have their origins in the
checkpointing techniques developed in the context of automatic differentiation
(AD). Because of this application context, these works have focused mainly on
the case of homogeneous chains, i.e. models consisting of a sequence of
identical blocks. In this context, it is possible to rely on dynamic
programming to find optimal solutions and even closed form formulas can be
derived to automatically find the activations to keep and the ones to delete.
From the complexity point of view, it has been shown that the problem is
NP-complete as soon as one considers graphs more general than chains in
\cite{naumann2008call}. In the re-materialization literature dedicated to DNN
training, we can distinguish approaches that focus on the case of sequential models
and those that consider more general graphs.

In the case of sequences, in~\cite{chen2016training}, the sequence on length $N$ 
is divided into $\sqrt{N}$ equal-length segments of length  $\sqrt{N}$, 
and only the input of each segment is
materialized during the forward phase.
This strategy is implemented in PyTorch in \texttt{torch.utils.checkpoint}.

Rotor~\cite{rotor-RR, rotor} provides optimal solutions in the case of fully
heterogeneous sequential models. Rotor is based on the evaluation of the
parameters of each block the sequence (computational and memory costs), on the
resolution of the re-materialization problem using dynamic programming and on
the implementation of the resulting computation sequence. Rotor is fast, but
it is limited to sequential models. It is one of the two ingredients (with
Checkmate~\cite{jain2020checkmate} described below) of the present contribution.


Other contributions target more general graphs. For example, the approach
described in~\cite{kumar2019efficient} is based on the computation of a
tree-width decomposition of the graph to determine the minimum computational
cost associated with the minimum possible memory footprint.
In~\cite{kusumoto2019graph}, an enumeration of subgraphs is required to design
efficient re-materialization strategies. In general, finding the evaluation
order of the graph that minimizes the memory consumption is a hard problem,
independently of any re-materialization strategy, as demonstrated
in~\cite{steinerolla}. The case of (non-optimal) dynamic re-materialization,
especially in the case where the input size in unknown in advance, has been
addressed in~\cite{kirisame2020dynamic,mimose}.
An important contribution in the case of general graphs
has been provided in Checkmate~\cite{jain2020checkmate}, which proposes an
Integer Linear Program (ILP) to find the optimal re-materialization sequence
for general graphs, which is consistent with the NP-Completeness results
of~\cite{naumann2008call}. An important limitation of Checkmate (see
Section~\ref{sec:models})
is the long solving time of the ILP solver, which limits its use to relatively small graphs.     
This paper addresses this
    limitation of Checkmate by proposing to combine it with Rotor.

\section{Rockmate}
\label{sec:models}

\subsection{Sketch of the Algorithm}
\label{sec:sketch}


As explained in Section~\ref{sec:intro}, the main idea of this paper is to
combine the ideas of (i) Checkmate, which finds good solutions in the case of
general graphs but is slow, and (ii) Rotor, which finds the optimal solution
only in the case of sequential networks, but is fast.

The GPT neural networks used as motivational example above is not completely sequential, but it can
be decomposed in a sequence of blocks, where each block contains
several operations. It is a typical example where, in order to use Rotor, it is necessary to
aggregate all the operations of the same block together. Rotor
therefore decides at the scale of the whole block whether to keep all
the data or to delete them all during the forward phase. Checkmate, on
the other hand, sees the whole graph describing the model and can
therefore decide, independently and at the level of each operation,
whether to keep its data or not.
 


The solution we propose is called Rockmate; a pseudo-code is provided
in Algorithm~\ref{alg:Rockmate} and explained below. The main idea is
to apply \textbf{Checkmate inside each block} and to apply
\textbf{Rotor on the complete sequence of blocks}. For this purpose,
it is necessary to obtain the complete graph of all operations of the
neural network, and to adapt both Checkmate and Rotor to this new
setting.



\begin{algorithm}[tb] \caption{Rockmate} \label{alg:Rockmate}
    \begin{algorithmic}[1]
        \STATE {\bfseries Input:} $module$, $input$, $M_{GPU}$
        \STATE $[blocks]$ = rk-GB($module$, $input$)
        \STATE $budgets$ = $[(M_{peak}, M_{save})]$ (quantized)
        \STATE $sols$ = []
        \FORALL {$b \in [blocks]$}
        \FORALL { $(M_{peak}, M_{save}) \in budgets$}
        \STATE \hspace{-0.15cm}$sols[b]$.add(rk-Checkmate($b$, $M_{peak}$, $M_{save}$))
        \ENDFOR
        \ENDFOR
        \STATE $Sequence$ = rk-Rotor($sols$, $M_{GPU}$)
       \STATE $rkMod$.forward = rk-Exec($Sequence.fwd$)
        \STATE $rkMod$.backward = rk-Exec($Sequence.bwd$)

        \STATE {\bfseries Output:} $rkMod$
    \end{algorithmic}
\end{algorithm}

The first phase is called \textbf{rk-GB} (for GraphBuilder). It
occurs on line 2 of Algorithm~\ref{alg:Rockmate} and is described in more
details in Section~\ref{sec:graphbuilder}. rk-GB takes as input a model expressed as
a PyTorch \texttt{nn.Module} and automatically 
(i) extracts the Directed Acyclic Graph
(DAG) of all the operations performed in the model, (ii) divides it
into a sequence of blocks and (iii) detects all the blocks which have
identical structures. For each unique block, the processing times of
all operations and the sizes of all intermediate data that are
produced by these operations are measured. These measurement (graphs
of each block, labeled with the execution times and memory footprints
of the produced data) contain all the necessary information to find
the re-materialization sequence. 

In the second phase of Rockmate (line 3-8 of
Algorithm~\ref{alg:Rockmate}), we consider each single block
independently. As we saw in the example in Figure~\ref{fig:checkmate_lin},
Rotor fails to compute very good re-materialization strategies because
it can only choose between two options: keep all or delete all
activations in the block. In Rockmate, we use a refined version of
Checkmate to generate a larger set of re-materialization
strategies. This refined version is denoted as \textbf{rk-Checkmate} and
described in Section~\ref{sec:rk_checkmate}.  

 A re-materialization strategy is characterized by (i) the memory peak
 during the execution of the block (either during forward or backward)
 and (ii) the total size of the internal activations of the block that
 are kept between the forward phase and the backward phase. The first
 one ensures that this strategy can be executed within a given memory
 limit. The second one allows the dynamic program to know how much
 memory will be left for the next blocks. The number of different
 options to consider is a parameter of Rockmate. We analyze its effect
 on performance in Section~\ref{sec:experiments} and show that
 quantizing each parameter into 20 different thresholds is enough to
 get good solutions in practice. This leads to at most 400 different
 strategies in total for each block. Since rk-Checkmate is applied at
 the level of a block (and not on the whole network), the
 corresponding graph is small enough that the runtime remains small,
 even for generating the whole family of strategies. Moreover, as
 rk-GB automatically detects identical blocks, rk-Checkmate is
 performed only on unique types of blocks (for instance, GPT2 models
 only involve five unique types of blocks). In practice, it takes less
 than 2 minutes to solve rk-Checkmate 400 times for a rk-block in GPT2,
 while it's impossible to solve the entire network with the ILP method
 because of its exponential complexity.

 The third phase of Rockmate (line 9 of Algorithm~\ref{alg:Rockmate},
 described in Section~\ref{sec:rk_DP}) is called \textbf{rk-Rotor} and computes the global
 re-materialization strategy. rk-Rotor features an adapted dynamic
 program of Rotor that, instead of having two solutions per block, can exploit the different
 re-materialization strategies computed during the second phase. The
 output of rk-Rotor therefore consists in a schedule which
 describes which block should be computed, in which order, and with
 which re-materialization strategy. If necessary, some blocks can be
 computed without keeping any data at all, and thus be recomputed later
 (possibly several times).
 
 Finally, the fourth phase (line 10 of Algorithm~\ref{alg:Rockmate},
 described in Section~\ref{sec:exec}) is called \textbf{rk-Exec}. It transforms this schedule into a new
 PyTorch \texttt{nn.Module}, which performs all the
 corresponding elementary operations in the correct order. The
 resulting module computes exactly the same gradients as the original
 version while respecting a global constraint on the memory usage of
 activations, at the cost of duplicating some computations.

\subsection{Phase 1: rk-GB, Graph Builder}

\label{sec:graphbuilder}
A typical training iteration of neural networks can be separated
as forward and backward phases. Both phases can be represented by
a data-flow graph. The computational graph
is explicit in TensorFlow, for which Checkmate was originally implemented.
In PyTorch, however, graphs need to be obtained by certain tools.
We developed a tool named rk-GraphBuilder (rk-GB)
which takes as input a \texttt{nn.Module} and
an example input for it, and builds the data-flow graph of the module.
Having an example input is necessary to inspect the time and memory cost
of all the operations used during forward and backward phases.

\paragraph{Obtaining the graph}
rk-GB does not require any modification or annotation of the module source code, 
instead it uses \texttt{torch.jit} to trace the forward execution of the
module on the example input. This function executes the forward
code and provides the list of all primitive operations used.
Based on this list, we build a forward graph where each node represents one assignment.
However, multiple variables may share the same memory space
due to \texttt{view} and \texttt{in-place} operations in PyTorch. 
Such variables would thus be kept or
removed together when performing re-materialization.
Therefore, rk-GB merges all the nodes sharing the same memory space
to obtain a simplified forward graph.
For a 12-layer GPT model, the number of nodes decreases from 934 to
185 after simplification. The simplified forward graph is further cut
around 1-separators: a node is a 1-separator if by removing it, we
obtain a disconnected graph (1 node to separate the graph).
This produces a sequence of blocks, as required by rk-Rotor.
For a 12-layer GPT model, this results in 26 blocks, where each
Transformer layer is separated to an Multi-Head Attention block
and an MLP block.

\paragraph{Identical blocks}
Afterwards, rk-GB goes through all the blocks to recognize \emph{identical
blocks}, \emph{i.e.} blocks whose computational graphs are the same.
Since rk-GB is deterministic, 
two blocks representing the same function share the same graph structure,
including the same topological ordering of nodes.
Following this ordering, rk-GB checks equivalency node by node.
A group of identical blocks can be measured and solved
together to improve the solving time.
Identifying identical blocks is an optimization 
of the Rockmate solving time but does not change
its solution quality. So even if some blocks are
wrongly declared as different by Rockmate, 
this would not change the memory gains or the computational overhead.
For a 12-layer GPT model, this procedure identifies only 5 
identical blocks from the 26 rk-blocks produced after separation.


\paragraph{CD\_graphs}
One underlying assumption in the original Checkmate graph model is
that each operation has exactly \emph{one} output data. However, when
several forward operations share the same input, the corresponding
backward operations contribute to the same data (by summing all the
contributions). This means that removing the result of one of these
backward operations has an impact on the other operations, which can
not be taken into account in the graph model of
Checkmate. Additionally, some elementary operations in PyTorch
actually create intermediate data (they are called
\texttt{saved\_tensors}), which can be deleted independently of the
output of the operation.

For these reasons, we introduce a new graph called
\texttt{CD\_graphs}, which contain two categories of nodes:
Computation and Data.  A \texttt{C\_node} represents an operation,
labeled with the time it takes and the temporary memory overhead
during execution.
A \texttt{D\_node} represents a data tensor stored in the memory.
A \texttt{D\_node} can be forgotten to free memory, and
restored by recomputing the corresponding \texttt{C\_node}s. An edge between
a \texttt{C\_node} and a \texttt{D\_node}
represents the execution dependency between
the operation and its output data tensors.
The benefits of considering such a \texttt{CD\_graph} is to enable
finer rematerializations,
such as releasing memory from a subset of outputs of one operation. 
The final product of rk-GB is a sequence of \texttt{CD\_graph}s.
More details about rk-GB can be found in
Section~\ref{sec:appendix:rk-gb} of the Appendix.

\subsection{Phase 2: rk-Checkmate, Options at Block Level}
\label{sec:rk_checkmate}

Given a \texttt{CD\_graph}, it is a non-trivial problem to find the
optimal execution schedule of all the operations within a given memory
limitation. To solve this problem, we use rk-Checkmate, an Integer
Linear Programming (ILP) adapted from
Checkmate~\cite{jain2020checkmate}. Just like Checkmate, rk-Checkmate
requires a topological order of all the operations, which is provided
by rk-GraphBuilder.  rk-Checkmate provides several improvements over
the original Checkmate formulation.

First, additional variables are introduced to represent the execution
of each \texttt{C\_node} separately from the memory allocation of each
\texttt{D\_node}. Constraints are also adapted to ensure that the
execution order follows the dependencies between computational nodes
and data nodes.  In the case where one operation generates multiple
outputs, there are multiple \texttt{D\_node}s depending on the same
\texttt{C\_node}. Deleting these outputs is considered separately in
rk-Checkmate, whereas they are grouped together in the Checkmate
formulation. For example, this improvement is useful for an operation
which produces two large outputs, each required by a different
operation: with rk-Checkmate, it is possible to delete the second
output before performing the operation that requires the first output,
which reduces the memory usage.


Second, rk-Checkmate takes into account the temporary memory usage of
all operations: because of temporary data allocated and deleted during the operation, 
the peak memory might be higher than the size of input and output.
Checkmate ignores this possibility, and
thus may produce solutions whose actual peak memory is higher than the
budget.

Finally, since rk-Checkmate is aware of the separation between forward
and backward phases, it is possible to include a constraint on the
memory usage when going from the forward to the backward phase. This
constraint expresses the limit $M_{save}$ on the size of the
activations which are kept in memory between both phases of a block
(and thus, during the execution of the following blocks).
This memory occupancy is necessary to control the overall
memory cost of all the blocks.
More details about rk-Checkmate can be found in
Section~\ref{sec:appendix:ilp} of the Appendix.


For each block, rk-Checkmate will be applied with different values for
the memory budgets $M_{peak}$ and $M_{save}$, as explained in
Section~\ref{sec:sketch}. We first compute the minimum and maximum
possible values for $M_{peak}$, by analyzing the memory usage of the
schedule which deletes activations as soon as possible, and of the
schedule which performs no recomputation, respectively. The number of
budgets is a hyperparameter of Rockmate whose effect is analyzed in
Section~\ref{sec:precision}. The values of $M_{peak}$ are evenly
spaced within $[min\_peak; max\_peak]$. Given one value
for $M_{peak}$, the values of $M_{save}$ are evenly spaced within
$[output\_size; M_{peak}]$. This ensures that all pairs $(M_{peak},
M_{save})$ given to rk-Checkmate are relevant. 
Note that different budgets may lead to the same optimal solution.
In practice, when we apply the number of $M_{peak}$ and $M_{save}$
as (20, 20) for GPT2, 
there are less than 30 unique solutions per rk-block.

Note that identical blocks are solved only once with the same budgets,
so that all identical blocks have the same set of block-level
execution options provided to rk-Rotor. 
However, rk-Rotor sees all of these blocks as different parts of the sequence, 
which just happen to have the same set of options
In the resulting sequence, each of these identical blocks may be 
executed with a different option in the output of rk-Rotor.

\subsection{Phase 3: rk-Rotor, Global Sequence Generation.}
\label{sec:rk_DP}

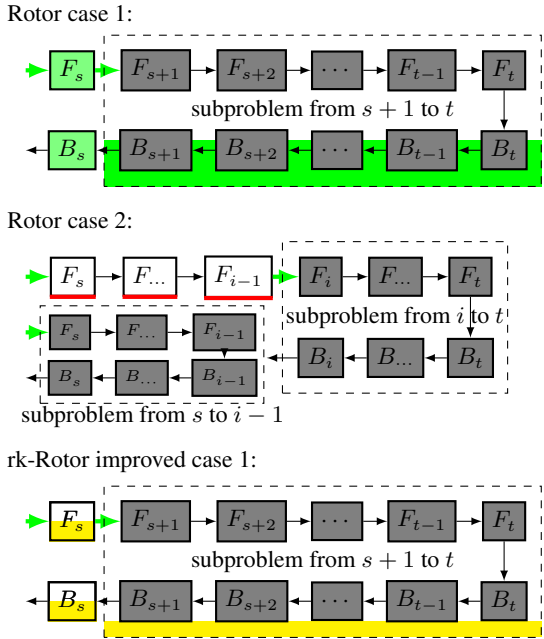
\begin{figure}
  \centering
\tikzset{
  op/.style={rectangle, thick, draw=black, font=\footnotesize, minimum height=.5cm},
  opfull/.style = {op, fill=green!50},
  fwopG/.style={op},
  bwopG/.style={fwopG},
  oprec/.style = {op, fill=black!50},
  saved/.style={color=green, ultra thick},
}
\newcommand{\fullnode}[2]{
  \node(#2) [opfull]{#1};
}
\newcommand{\emptynode}[2]{
  \node(#2) [op]{#1}; \draw[ultra thick, red] (#2.south west) -- (#2.south east);
}
\newcommand{\halfemptynode}[2]{
  \node(#2) [op]{#1};\path[fill=yellow] (#2.south west) rectangle (#2.east);
  \node(#2) [op]{#1};
}
\newcommand{\recnode}[2]{
  \node(#2) [oprec]{#1};
}

\begin{tikzpicture}[>=latex, every node/.style={font=\footnotesize}]

  \begin{scope}

    \node[anchor=south west] at (0, 2) {Rotor case 1:};
    
    \matrix[matrix anchor=south west,row sep=0.5cm,column sep=0.3cm,ampersand replacement=\&] {
      \node (finput)  [] {};	\&
      \fullnode{$\fop{s}{}$}{f_1};	\&
      \recnode{$\fop{s+1}{}$}{f_2} \&
      \recnode{$\fop{s+2}{}$}{f_3} \&
      \recnode{$\cdots$}{f_4} \&
      \recnode{$\fop{t-1}{}$}{f_5} \&
      \recnode{$\fop{t}{}$}{f_6} \&
      \node (foutput) [] {};\&
      \\
      \node (boutput) [] {};	\&
      \fullnode{$\bop{s}{}$}{b_1};	\&
      \recnode{$\bop{s+1}{}$}{b_2} \&
      \recnode{$\bop{s+2}{}$}{b_3} \&
      \recnode{$\cdots$}{b_4} \&
      \recnode{$\bop{t-1}{}$}{b_5} \&
      \recnode{$\bop{t}{}$}{b_6} \&
      \node (binput) [] {};	\&
      \\
    };
    
    \path[->]
    (finput) edge[saved](f_1)
    (f_1) edge[saved] (f_2)
    (f_2) edge (f_3)
    (f_3) edge (f_4)
    (f_4) edge (f_5)
    (f_5) edge  (f_6)
    (f_6) edge  (b_6);
    \path[->]
    (b_6) edge (b_5)
    (b_5) edge (b_4)
    (b_4) edge(b_3)
    (b_3) edge (b_2)
    (b_2) edge (b_1)
    (b_1) edge (boutput)
    ;

    \draw[dashed] ($(b_2.south west) + (-0.2, -0.2)$) rectangle ($(f_6.north east) + (0.2, 0.2)$) node[midway]{subproblem from $s+1$ to $t$};

    \begin{scope}[on background layer]
      \fill[green] ($(b_2.south west) + (-0.2, -0.2)$) rectangle ($(f_6.north east) + (0.2, -1.2)$);
    \end{scope}
    
  \end{scope}

  \begin{scope}[yshift=-2.75cm]

    \node[anchor=south west] at (0, 2) {Rotor case 2:};

    \matrix[matrix anchor=south west,row sep=0.5cm,column sep=0.3cm,ampersand replacement=\&] {
      \node (finput)  [] {};	\&
      \emptynode{$\fop{s}{}$}{f_1};	\&
      \emptynode{$\fop{\cdots}{}$}{f_2} \&
      \emptynode{$\fop{i-1}{}$}{f_3} \&
      \recnode{$\fop{i}{}$}{f_4} \&
      \recnode{$\fop{\cdots}{}$}{f_5} \&
      \recnode{$\fop{t}{}$}{f_6} \&
      \node (foutput) [] {};\&
      \\
      \&\&\&\node (b_3) [text width=.5cm]{}; \&
      \recnode{$\bop{i}{}$}{b_4} \&
      \recnode{$\bop{\cdots}{}$}{b_5} \&
      \recnode{$\bop{t}{}$}{b_6} \&
      \node (binput) [] {};	\&
      \\
    };
    
    \path[->]
    (finput) edge[saved](f_1)
    (f_1) edge (f_2)
    (f_2) edge (f_3)
    (f_3) edge[saved] (f_4)
    (f_4) edge (f_5)
    (f_5) edge  (f_6)
    (f_6) edge  (b_6);
    \path[->]
    (b_6) edge (b_5)
    (b_5) edge (b_4)
    (b_4) edge(b_3)
    ;

    \draw[dashed] ($(b_4.south west) + (-0.2, -0.2)$) rectangle ($(f_6.north east) + (0.2, 0.2)$) node[midway]{subproblem from $i$ to $t$};
    
    \matrix[matrix anchor=south west, row sep=0.1cm,column sep=0.3cm,ampersand replacement=\&] at (0, -0.25) {
      \node (finput)  [] {};	\&
      \node[oprec, minimum height=0.25cm, font=\tiny] (f_1) {$\fop{s}{}$};	\&
      \node[oprec, minimum height=0.25cm, font=\tiny] (f_2) {$\fop{\cdots}{}$};	\&
      \node[oprec, minimum height=0.25cm, font=\tiny] (f_3) {$\fop{i-1}{}$};	\&
      \\
      \node (boutput) [] {};	\&
      \node[oprec, minimum height=0.25cm, font=\tiny] (b_1) {$\bop{s}{}$};	\&
      \node[oprec, minimum height=0.25cm, font=\tiny] (b_2) {$\bop{\cdots}{}$};	\&
      \node[oprec, minimum height=0.25cm, font=\tiny] (b_3) {$\bop{i-1}{}$};	\&
      \\
    };

        \path[->]
    (finput) edge[saved](f_1)
    (f_1) edge (f_2)
    (f_2) edge (f_3)
    (f_3) edge  (b_3);
    \path[->]
    (b_3) edge (b_2)
    (b_2) edge (b_1)
    (b_1) edge (boutput)
    ;

    \draw[dashed] ($(b_1.south west) + (-0.1, -0.1)$) rectangle ($(f_3.north east) + (0.1, 0.1)$);
    \path ($(b_1.south west) + (-0.1, -0.1)$) -- ($(b_3.south east) + (0.1, 0.1)$) node[midway, below]{subproblem from $s$ to $i-1$};

\end{scope}

  \begin{scope}[yshift=-6cm]
    
    \node[anchor=south west] at (0, 2) {rk-Rotor improved case 1:};

    \matrix[matrix anchor=south west,row sep=0.5cm,column sep=0.3cm,ampersand replacement=\&] {
      \node (finput)  [] {};	\&
      \halfemptynode{$\fop{s}{}$}{f_1};	\&
      \recnode{$\fop{s+1}{}$}{f_2} \&
      \recnode{$\fop{s+2}{}$}{f_3} \&
      \recnode{$\cdots$}{f_4} \&
      \recnode{$\fop{t-1}{}$}{f_5} \&
      \recnode{$\fop{t}{}$}{f_6} \&
      \node (foutput) [] {};\&
      \\
      \node (boutput) [] {};	\&
      \halfemptynode{$\bop{s}{}$}{b_1};	\&
      \recnode{$\bop{s+1}{}$}{b_2} \&
      \recnode{$\bop{s+2}{}$}{b_3} \&
      \recnode{$\cdots$}{b_4} \&
      \recnode{$\bop{t-1}{}$}{b_5} \&
      \recnode{$\bop{t}{}$}{b_6} \&
      \node (binput) [] {};	\&
      \\
    };
    
    \path[->]
    (finput) edge[saved](f_1)
    (f_1) edge[saved] (f_2)
    (f_2) edge (f_3)
    (f_3) edge (f_4)
    (f_4) edge (f_5)
    (f_5) edge  (f_6)
    (f_6) edge  (b_6);
    \path[->]
    (b_6) edge (b_5)
    (b_5) edge (b_4)
    (b_4) edge(b_3)
    (b_3) edge (b_2)
    (b_2) edge (b_1)
    (b_1) edge (boutput)
    ;

    \draw[dashed] ($(b_2.south west) + (-0.2, -0.2)$) rectangle ($(f_6.north east) + (0.2, 0.2)$) node[midway]{subproblem from $s+1$ to $t$};
    
    \begin{scope}[on background layer]
      \fill[yellow] ($(b_2.south west) + (-0.2, -0.2)$) rectangle ($(f_6.north east) + (0.2, -1.6)$);
    \end{scope}
  \end{scope}

\end{tikzpicture}
  \caption{Diagram representing the different cases for the dynamic
    program. Green arrows represent materialized activations. Green,
    yellow and red blocks represent internal activations to the
    blocks, which are respectively completely, partially, or not
    saved. Colored backgrounds on the subproblems represent 
    how much memory is occupied by these activations.}
  \label{fig:rk-rotor.cases}
\end{figure}

\paragraph{Principle of Rotor}
The main idea of the dynamic programming algorithm of Rotor is as
follows. An optimal solution for the forward-backward computation from
block $s$ to $t$ with memory $m$ can be of two different types: either
the first block $s$ is computed only once, or more than once. In the
first case, the computation starts with computing block $s$ and
keeping all intermediate data, and continues with an optimal solution
for blocks $s+1$ to $t$ (with less memory available). In the second
case, the computation starts with computing blocks $s$ to $s+i$ for
some $i$, stores the result of $s+i$, continues with an optimal
solution for blocks $s+i$ to $t$, and finally recomputes from $s$ to
$s+i$ with an optimal solution for this part. 
Note that no intermediate data is saved for blocks $s$ to $s+i$.
An illustration of each
case is represented on Figure~\ref{fig:rk-rotor.cases}.

In each case, the subproblems that need to be solved have a smaller
value of $t-s$. Assuming that the solutions to these smaller problems
are known, the algorithm can make the choice which leads to the
smallest overhead among all valid choices, \emph{ie} those for which
the memory usage is not higher than the budget $m$. We can thus
iteratively compute optimal solutions until we find the solution for
the complete model.

\paragraph{rk-Rotor}
In the Rockmate context, we have several different options for the
first case: we can choose to keep more or less intermediate data for
the first block $s$. Each of these options leads to a different memory
usage for storing the intermediate data and for computing the backward
operation. There is thus a larger set of choices to choose from, but
the main idea is still there: assuming that solutions to all smaller
problems are known, we can select the option that yields the lowest
overhead among all options which respect the memory budget. This
improved case is represented at the bottom of Figure~\ref{fig:rk-rotor.cases}.

\paragraph{Complexity analysis}
With a model that contains $L$ blocks, and a memory of size $M$, the
Rotor algorithm has a complexity in $O(L^3M)$: for each value of $s$,
$t$ and $m$, there are $O(L)$ choices to consider. In the Rockmate
case, with $B$ budget options, the dynamic programming algorithm
considers $O(L+B)$ choices at each step, and thus has a complexity in
$O(L^2M(L+B))$. More details about the rk-Rotor algorithm can be found
in Section~\ref{sec:appendix:rk-rotor} of the Appendix.

\paragraph{Sub-optimality of the solution}
Although both rk-Checkmate and rk-Rotor obtain optimal solutions
for the given sub-tasks, the final Rockmate solution is not always optimal
on the overall network. Two reasons can lead to sub-optimality
in Rockmate: (i) since the number of memory budgets is finite, only
a limited number of execution schedules are produced by rk-Checkmate.
(ii) in rk-Rotor, intermediate data is only used to improve the
execution time of the backward phase. However, if the forward phase of
a block is executed several times, it might be beneficial to save some
intermediate tensor on the first pass, and use it to compute the
output faster on subsequent passes. This possibility is not considered
in rk-Rotor: forward passes in case 2 do not save
intermediate data.

\subsection{Phase 4: rk-Exec}
\label{sec:exec}



Rockmate creates a PyTorch \texttt{nn.Module} that performs a forward-backward 
computation based on the optimal schedule solved by the algorithm described above. 
The execution of the forward phase is based on the Python code obtained via
\texttt{jit.trace}. 
For backward, the PyTorch autograd engine stores 
the ``computational graph'' during the forward phase, which allows backward 
computation from the output back to the input. 
For the sake of clarity, we call this an \emph{autograd graph}.
In Rockmate, we detach the operations during the forward phase, 
so that the full network is represented as many small autograd graphs. 
This allows the backward operations to be performed separately, thus
deletions of tensors can be easily inserted between two backward operations.
Specifically, rk-Exec creates one autograd graph for each \texttt{C\_node}
defined in Section~\ref{sec:graphbuilder}. Every autograd graph
contains all the operation which create tensors sharing the same memory space,
such as \texttt{torch.Tensor.view} operations. 
During the backward phase, the gradient of each tensor is
automatically supplied to the previous backward function. 
Furthermore, the recomputation of the operations have
to be performed in different ways, so that the existing autograd graph
will not be rebuilt. 
Also, when one operation will be recomputed before the backward,
the execution will be with \texttt{torch.no\_grad} mode so that
\textit{saved\_tensors} will not be created.
More details about the rk-Rotor algorithm can be found
in Section~\ref{sec:appendix:rk-exec} of the Appendix.

\section{Experiments}
\label{sec:experiments}

\subsection{Experimental Settings}

All experiments presented in this paper are performed using Python 3.9.12 and PyTorch 1.13.0. Rockmate, Rotor, and Checkmate compute their solutions on a 40-core Intel Xeon Gold 6148, while training is performed on an Nvidia Tesla V100 GPU with 15.75 GB of memory. For comparison with Rotor on ResNet and GPT2, both networks are implemented as a \texttt{nn.Sequential} module of PyTorch. For RegNet, we do not provide a comparison with Rotor due to the lack of a Rotor-compatible implementation. 
All experiments use the version of Rockmate available at:
\url{https://github.com/topal-team/rockmate/releases/tag/v1.0}.

Rockmate is used to reduce peak memory usage due to storage of activations. We measure and control the memory footprint of activations during the experiments. The memory used by the model parameters is excluded from our peak memory budget, as it remains constant during training. In practice, we adjust the maximum peak memory for activations by subtracting the size of the model from the total memory.



\subsection{Precision}
\label{sec:precision}
\begin{figure}          
    \centering
    \includegraphics[width=\linewidth]{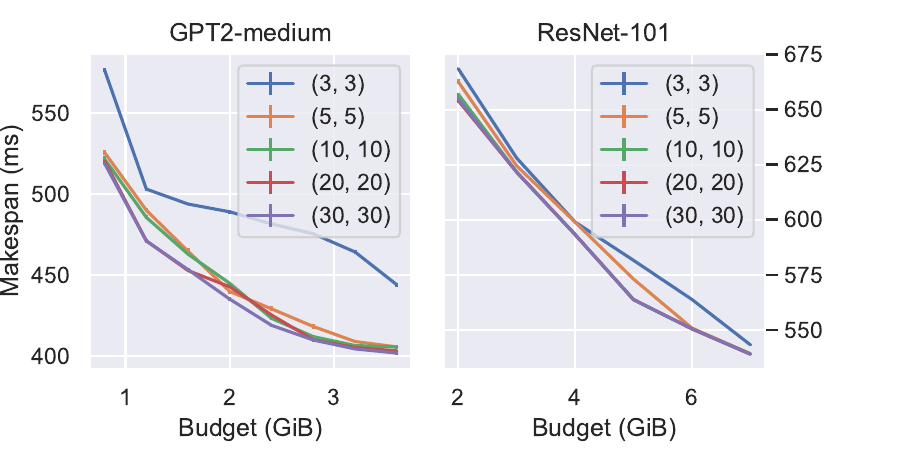}
    \caption{Experiments on GPT2-medium and ResNet-101. Number
        of $(M_{peak}, M_{save})$ budgets are chosen from (3,3) to (30,30).
        Makespan is the time of one training iteration (including forward and backward).
    }
    \label{fig:options}
\end{figure}

As discussed in Section~\ref{sec:rk_checkmate}, the same number of budget options are 
used to solve each block in rk-Checkmate. Increasing the number of options increases 
the time to run Rockmate, since it is directly related to the number of times rk-Checkmate 
is called for each block. However, it provides finer re-materialization 
strategies for rk-Rotor. Figure~\ref{fig:options} shows how 
the number of budget options affects the quality of the Rockmate solution. 
Budget options range from (3,3) to (30,30) for GPT2-medium and ResNet-101. 
Overall, increasing the number of budget options improves Rockmate performance up to a point.  
Specifically, the improvement in the Rockmate solution is stronger on GPT2-medium 
than on ResNet when more budget options are allowed. 
This is because a GPT2 block contains more complicated structures (more nodes in block), 
while a ResNet block is too small to apply sophisticated re-materialization strategies. 
In the following experiments, we use the number of $(M_{peak}, M_{save}) = (20,20)$ for rk-Checkmate.

\subsection{Efficiency}
\label{suc:efficiency}

\begin{figure}
    \centering
    \includegraphics[width=\linewidth]{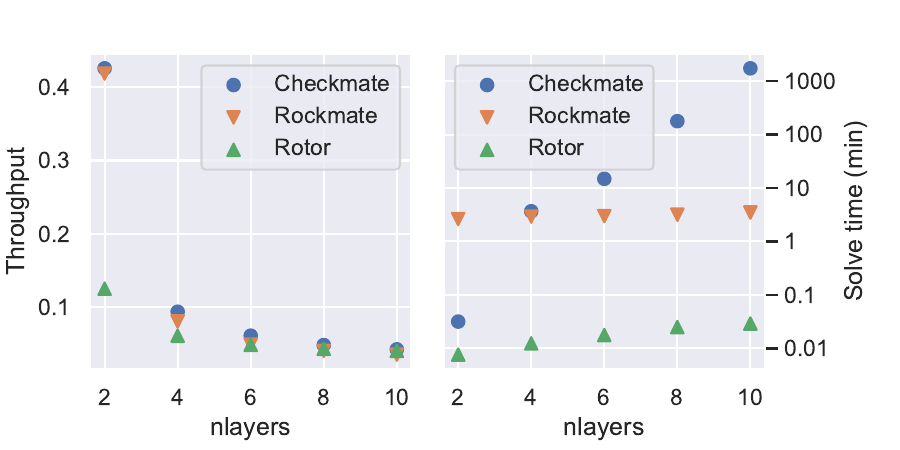}
    \caption{Experiments on solving GPT2 with 2-10 Transformer blocks
        with given budgets. Simulation solving time is in log scale.
        Throughput is defined as the number of samples processed per time unit (ms).
    }
    \label{fig:checkmate_log}
\end{figure}

In Figure~\ref{fig:checkmate_log}, we compare the solving time and performance of Rockmate, 
Checkmate~\cite{jain2020checkmate}, and Rotor~\cite{beaumont2020optimal} on GPT networks
with 2 to 10 Transformer blocks.
For each network, we choose the smallest memory budget
for which we obtain feasible solutions.
Checkmate is solved until convergence. Rockmate achieves very similar throughput to Checkmate, 
while Rotor can be significantly worse if the network is not deep enough. 
Since Rotor has only the option to recompute a whole block, it is more effective when the network is deep.

The time to solution of Rockmate remains nearly the same as the network gets deeper. 
The processing time of Rockmate consists of three parts: 
1. inspection time during graph building; 
2. rk-Checkmate processing time; 
3. rk-Rotor processing time. 
As described in Section~\ref{sec:graphbuilder}, 
rk-GB automatically detects identical blocks. 
Only one inspection is performed for a class of identical blocks, 
and they are solved by the same rk-Checkmate models. 
Therefore, the inspection time and the rk-Checkmate time remain 
the same when the number of identical transformer blocks increases. 
The rk-Rotor solving time is similar to Rotor's, 
which is much faster than the total Rockmate solving time.    

The complexity of Checkmate's ILP model grows exponentially with the size of the network, 
making it infeasible on modern neural networks with thousands of nodes. 
In Figure~\ref{fig:checkmate_log} we compare the solving time and 
performance of Checkmate and Rockmate on GPT2 with 2 to 10 Transformer blocks. 
The memory budgets are chosen as the minimum achievable budgets for both models. 
The solving time of Checkmate is exponential in the number of blocks, 
exceeding 30 hours on the 10-blocks GPT2. On the other hand, 
the solution time of Rockmate remains almost constant because 
the same rk-Checkmate models are applied to all identical 
Transformer blocks. Despite the significant difference in 
solution time, Rockmate achieves similar or better overhead than Checkmate within the same budget.

\subsection{Performance}

\begin{figure}
    \centering
    \includegraphics[width=\linewidth]{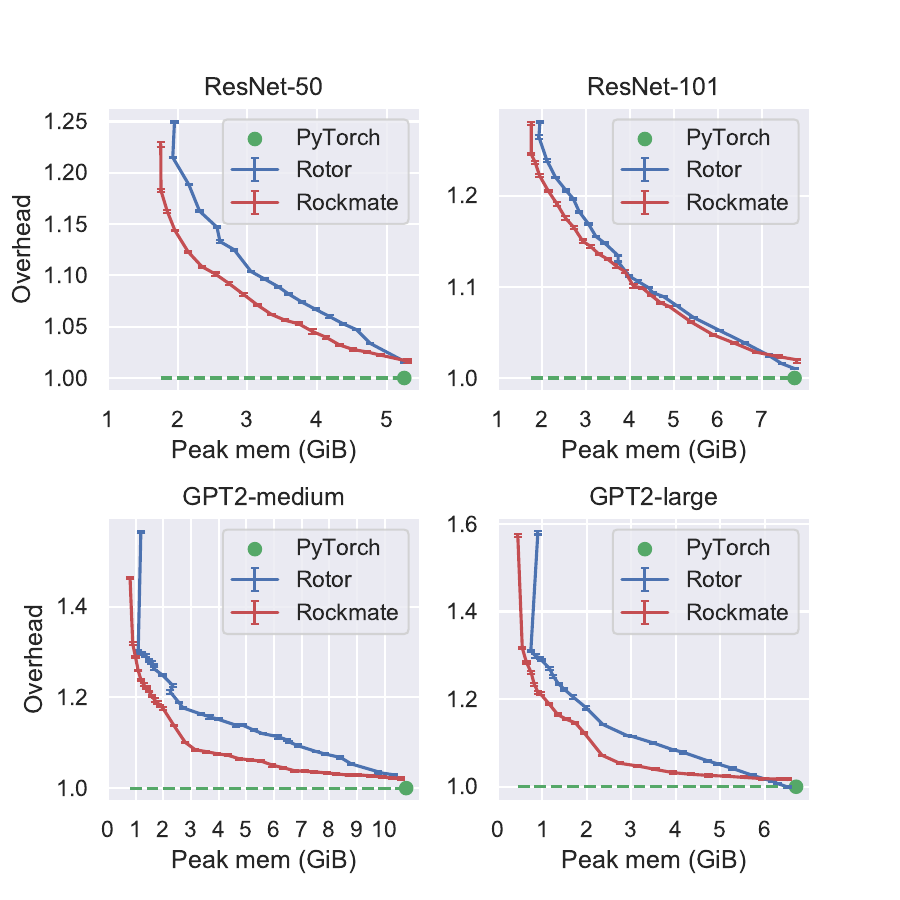}
    \caption{Computational overhead versus peak memory usage 
    on different networks. Within the same memory usage, 
    Rockmate is faster than Rotor in most cases. On GPT-like
    models, Rockmate significantly outperforms Rotor. 
    }
    \label{fig:compare_rotor}
\end{figure}

We compare Rockmate with Rotor on ResNet and GPT2 models over a range of memory budgets. 
Figure~\ref{fig:compare_rotor} shows the computational overhead in terms of peak memory 
usage during the forward-backward computations. 
For the same memory peak, Rockmate has a lower overhead than Rotor in most cases. 
For ResNet models, Rockmate does not show a significant improvement over Rotor, 
especially when the neural networks are deep enough, 
in which cases Rotor has more re-materialization options. 
On the other hand, Rockmate shows much better performance 
than Rotor on GPT2 networks. For GPT2-large it is noteworthy 
that Rockmate saves 50\% memory by introducing only 5\% overhead, 
while Rotor has more than 10\% overhead for the same budget. 
In addition, Rockmate allows training with a smaller memory budget. To train GPT2-large, Rotor requires at least 720 MB memory budget, while Rockmate only requires 440 MB.

The reason why Rockmate significantly outperforms Rotor is that there are "cheap" operations inside a Transformer block, such as \texttt{dropout} and \texttt{gelu}. The tensors generated by these operations consume a lot of memory, but there is almost no cost to recompute these operations. Because Rotor rematerializes one block at a time, it cannot take advantage of the "cheap" operations to optimize performance. Rockmate works particularly well on models with a sequential-like structure, where each part contains a complicated structure. 

\begin{figure}
    \centering
    \includegraphics[width=\linewidth]{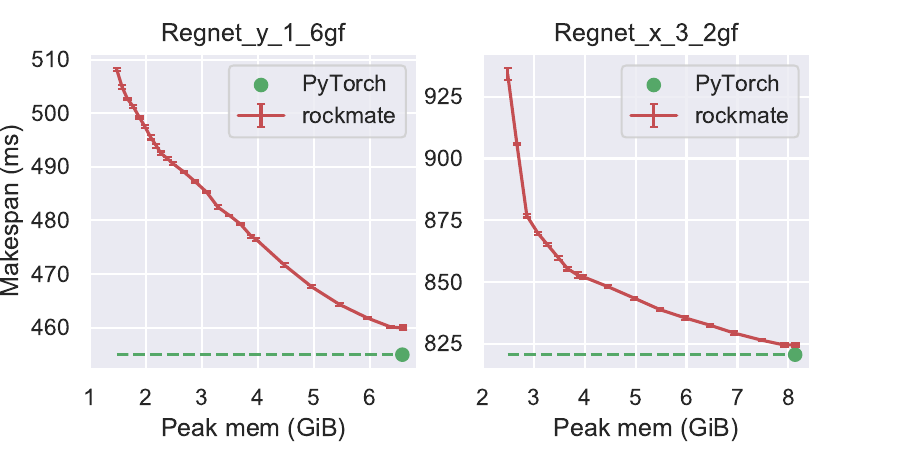}
    \caption{RegNet imported from \texttt{torchvision} is tested by Rockmate.
    Makespan is the time of one training iteration (including forward and backward).
    }
    \label{fig:regnet}
\end{figure}

While Rotor can only handle \texttt{nn.Sequential}-type models as input, 
Rockmate can be applied to more general types of neural networks.  
In Figure~\ref{fig:regnet} we show the results of using Rockmate directly on the RegNet model 
imported directly from \texttt{torchvision}, 
whereas using it in Rotor would require to rewrite the code to highlight the sequential part.
Although performance may vary depending on the structure of the neural networks, 
Rockmate can be used to save memory on general PyTorch models.

\section{Conclusion and Perspectives}
\label{discussion}

In this paper, we propose Rockmate, a fully automatic tool that takes as input a
PyTorch model in the form of a \texttt{nn.Module} and a memory limit for 
activations and automatically generates another \texttt{nn.Module}, perfectly
equivalent from the numerical point of view, but that fulfills the memory limit 
for activations at the cost of a small computational overhead. Through 
experiments on various models, we show that the computation time of the resulting 
\texttt{nn.Module} is negligible in practice and that the computational overhead 
is acceptable, even for drastic reductions in memory footprint. Rockmate is 
therefore a tool that can transparently allow increasing model size, data 
resolution and batch size without having to upgrade GPUs. This work opens 
several new scientific questions. First, Rockmate is very efficient for graphs 
that can be written as a sequence of blocks, which corresponds to numerous 
models in practice but not to all of them, which raises the question of its 
extension to any type of graph. Then, the combination of Rockmate with data 
parallelism is trivial, but the question of finding a partition of the model 
adapted to model parallelism that balances well the computational load and the 
memory footprint on the different nodes is also an open problem. 

\paragraph{Acknowledgements}
This work has been partially funded by the Inria DFKI ENGAGE Project: \emph{nExt geNeration computinG environments for Artificial intelliGEnce}.

\newpage
\newpage

\bibliography{sample.bib} \bibliographystyle{icml2023}

\newcommand\jit{\textit{torch.jit.trace }}

\newpage \appendix \onecolumn

\newcommand\ie{\textit{i.e. }}
\newcommand{\ti}{\textit}
\newcommand\Cn{\textit{C\_node }}
\newcommand\Dn{\textit{D\_node }}
\newcommand\fw{\textit{forward }}
\newcommand\bw{\textit{backward }}

\section{rk-GraphBuilder}
\label{sec:appendix:rk-gb}

\subsection{Outline}

We have developed the Rockmate Graph Builder (rk-GB) as a tool for producing the graphs required for Rockmate. Although it was developed for this specific purpose, it can also be used independently. Given a \textit{torch.nn.Module} and a given input for that module, our goal is to generate a graph showing all the operations that occur during the \textit{forward} and \textit{backward} of the module on that given input. Having a specific input is important for both the resolution of if statements in the forward code and the inspection of each operation. Indeed, since our goal is to identify recomputations, we have to know the time spent and the memory footprint for each operation, that both depend on the input.

rk-GB relies on \jit to trace the execution of the 
\textit{forward} module on the input. This function runs the 
\textit{forward} code and returns the list of all performed operations. 
Based on this list, we build the \textit{forward} graph and transform 
it through several steps to generate what is needed by Rockmate.
Note that currently rk-GB may fail due to limitations of 
\textit{torch.jit.trace}, for instance it does not work properly 
on the GPT model from \textit{HuggingFace}. 
However, PyTorch 2.0 was recently announced with a new 
way to capture graphs: \textit{TorchDynamo}. 
According to its authors, \textit{TorchDynamo} is expected to work 
with almost all modules. 
We plan to try changing from \textit{jit.trace} to \textit{TorchDynamo}.

rk-GB consists of five steps. First, we build the forward graph based on 
\textit{jit.trace}, also collecting some information about each node. 
Then we \textit{simplify} the graph. This part significantly reduces the 
number of nodes, which is a nice feature for the ILP used in rk-Checkmate. 
This is more than just an optimization, it's a requirement for 
correct re-materialization. Next, we split the 
simplified forward graph into blocks using the 1-separator\footnote{a node is a 1-separator if by removing it, we
obtain a disconnected graph (1 node to separate the graph)} list.
This produces a sequence of forward graphs, which we refer to as 
\textit{blocks}. Rockmate will process each block with rk-Checkmate, 
and then process the whole chain with rk-Rotor. At this point, similar 
blocks are detected to avoid solving the same problem multiple times. 
Finally, for each unique forward block, we build the forward + backward 
graph and monitor the time and memory usage of each node during this step.

\subsection{The forward graph}

First, we call \textit{torch.jit.trace\_module} to get the \textit{forward} 
code of any given \textit{torch.nn.Module}. Specifically, we use the 
\textit{code\_with\_constants} attribute of the output object, which 
is a code string of the assignments made during the \textit{forward} phase. 
In Rockmate, a list of assignments is required, where each line consists 
of exactly one target and one operation. 

Therefore,\\
\texttt{a,b = torch.chunk(torch.relu(M),2)} \\ becomes \\
\texttt{fv\_1 = torch.relu(M) ; \\
    fv\_2 = torch.chunk(fv\_1,2) ; \\
    a = fv\_2[0] ; b = fv\_2[1]} \\

We also need to inline submodules code, so that \\
\texttt{b = self.layer1(input)  ; \\
    output = self.layer2(b)} \\
becomes\\
\texttt{
    \# layer1 part : \\
    fv1 = torch.linear(input, ...) ; \\
    fv2 = torch.relu(fv1) ; \\
    fv3 = torch.linear(fv2, ...) ; \\
    \# layer2 part : \\
    fv4 = torch.dropout(fv3,0.1) \\
    output = torch.layer\_norm(fv4,...)
} 

We can access the code of the submodules through the object returned by \textit{jit.trace\_module} with : \\
\textit{\textless jit\_output\textgreater.\textless submodule name\textgreater.code} \\ 
We assign a unique number to each target to avoid name collisions when building the submodule code.

\begin{figure}[hbtp]
    \caption{
        Forward graph of a GPT2 with 2 transformer blocks. \\
        All the figures are generated using rk-GB \textit{print\_graph}
        function, which relies on Graphviz.
    }
    \includegraphics[width=0.65\textwidth]{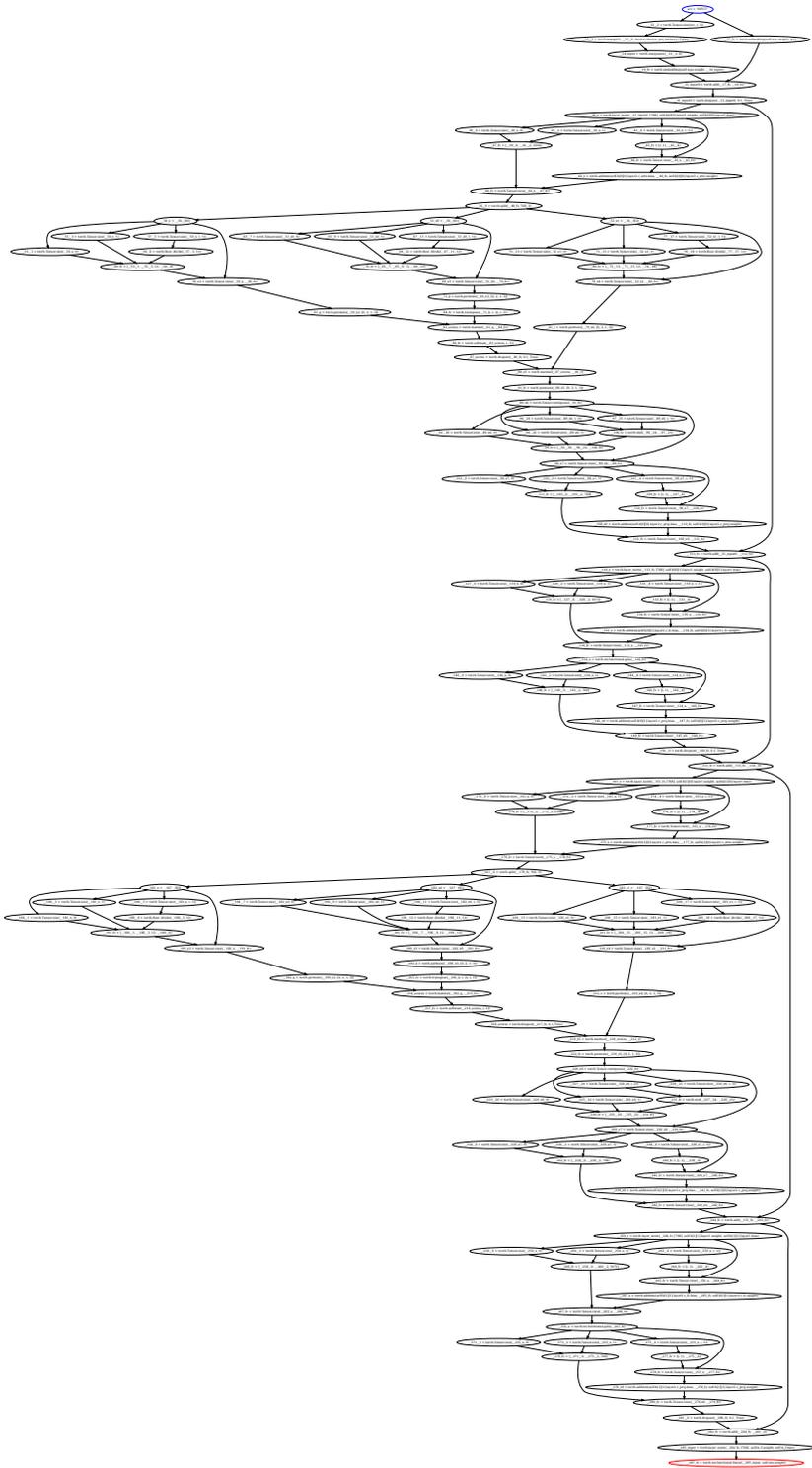}
\end{figure}

\subsection{Simplification}

In Rockmate, we want to perform both forward and backward operations on nodes, with the ability to delete and recompute tensors. Both view and in-place operations share their \textit{data} with their input, so that deleting a node without also deleting its views is not allowed. All nodes related to the same \textit{data} must be merged. After the simplification part, each node consists of exactly one primary allocation, which creates a data item, and some secondary allocations associated with that data, that do not allocate new memory.

First, we need to run each node to analyze it, 
since the name of the function might not be enough to determine whether the operation creates a new data item or not. For example, the function \textit{torch.contiguous} may create a new \textit{data item} depending on whether the input \textit{data} is contiguously stored in memory or not. Another common example is \textit{torch.reshape}, which is a visualization function if and only if the input and the requested shape have compatible strides. A robust way to check if two tensors refer to the same data is to compare their attribute \textit{data\_ptr}. To do this, we must create each variable, determine its type, and, if it is a tensor, check whether its \textit{data\_ptr} is new or not.

\paragraph*{Analysis of each node}

Note that we cannot analyze all of the code at once, as this would result in all intermediate activations being stored in memory, which is inconsistent with Rockmate's goal of using as little memory as possible. We will analyze each node separately, without storing any tensor. We proceed as follows
\begin{itemize}
    \item First, we randomly generate its inputs. Since we traverse the nodes in a topological order, the inputs have already been analyzed. For the reasons mentioned earlier, the tensors have not been stored, but their data type and shape have been recorded. Therefore, we can randomly generate the inputs based on this information. . For example, as mentioned earlier, the behavior of \textit{torch.reshape} depends on whether the input and the required shape have compatible strides. Therefore, to generate the inputs correctly, we regenerate the \textit{data} and perform all view operations on it before analyzing the node.
\item Then we run the code to analyze, which consists of an assignment, 
so that we get a value. If it is a \textit{torch.Size} (or something similar), 
we store it and mark it as a node of type \textit{size}. 
Otherwise it consists of either a tensor or a list of tensors. 
In both cases it has a \textit{data\_ptr}. First we check if the value already exists in the local directory, in which case it is an in-place operation. Consider the following example: \\
          \texttt{
              A = torch.linear(M,...) ; \\
              A += M
          }\\
It appears in the forward graph as \\
          \texttt{
              A = torch.linear(M,...) ; \\
              fv1 = torch.Tensor.add\_(A,M)
          }\\
When analyzing the node of \texttt{fv1}, we must recognize that it is the result of an in-place operation on \texttt{A}.           In this case, \texttt{fv1} refers to the same Python object as \texttt{A}. So to detect an in-place operation, we compare the address of the object \texttt{fv1} with its inputs (using the Python keyword \textit{is}). If we find a match, we mark the node as \textit{in-place} and record the name of its \textit{data owner} (in the example above, the \textit{data owner} of \texttt{fv1} is \texttt{A}). Otherwise, we check if the value shares its \textit{data\_ptr} with one of its inputs, in which case it is a node of type \textit{view}, and we record the name of its \textit{data owner}.           Finally, by default, the tensor is an original data and its \textit{data owner} is itself. We call this case by default the operation \textit{real}.
\item Finally, we store information such as the type and shape of the data so that we can randomly regenerate data when analyzing upcoming nodes.           Note that we also need to store the \textit{requires\_grad} information. It is essential for building the \textit{backward} graph, and we need to use it when regenerating tensors.
\end{itemize}

\paragraph*{Simplification process}

Once the analysis has been performed, the simplification of the graph can be started. The simplification is done in three steps:
\begin{itemize}
    \item
First, we process so-called \textit{cheap} operations, such as \textit{list} and \textit{tuple} constructors.           In contrast to the following simplification steps, here we insert the code to be simplified directly into the user code, for example\\
          \texttt{
              fv1 = [s1,s2] ; \\
              B = torch.reshape(A,fv1)
          }\\
          becomes \\
          \texttt{
              B = torch.reshape(A, [s1,s2])
          } \\
          Simplifying the \textit{list} and \textit{tuple} constructors is 
          mandatory to ensure each node represents only one tensor, it also 
          makes the code easier to read.
          \textit{cheap} operations also include \textit{torch.add}, \textit{sub}, \textit{mul}, and \textit{div}. This is due to the way \textit{autograd} handles intermediate results.           Normally, nested operations (neither primary nor inline) create intermediate variables that are stored in \textit{grad\_fn}. Therefore, we repeat the submodules until we reach primary operations.           But these specific operations do not create intermediate data in \textit{grad\_fn} (because we do not need them during \ti{backward}). Therefore, it is preferable not to inline them, as otherwise they would explicitly create intermediate variables and thus use more memory. Note that we could duplicate the code: \\
                    \texttt{
              A = torch.add(B, C) ; \\
              D = torch.linear(A,...) ; \\
              E = torch.relu(A)
          } \\
          becomes \\
          \texttt{
              D = torch.linear(\\
              \phantom{D = torch}torch.add(B, C),...) ; \\
              E = torch.relu(torch.add(B, C))
          } \\
Since these operations are fast enough (\textit{e.g.} compared to \textit{torch.matmul}), it will not take too much time. However, these simplifications represent a trade-off between the number of intermediate variables and the number of dependencies.           In the example above, \textit{D} now depends on both \textit{B} and \textit{C}. Therefore, it is optional to consider \textit{Add}, \textit{Sub}, \textit{Mul} and \textit{Div} as cheap operations.

\item The second step in the simplification process concerns nodes of type \textit{size}. Since these operations do not create data, we move them as secondary assignments of the node they refer to, which we call the \textit{body\_code}. To avoid creating new dependencies, for example in the case where a node depends only on the shape of a tensor and not on the tensor itself:\\
          \texttt{
              s = torch.Tensor.size(A,0) ; \\
              ... \\
              C = torch.reshape(B, [s])
          } \\

In this example, we are not yet sure whether \texttt{C} depends directly on 
\texttt{A}, but merging \texttt{s} into \texttt{A} would actually enforce 
the dependency. That would be wrong because it would cause 
Rockmate to conclude that \texttt{A} cannot be forgotten before computing 
\texttt{C}. To avoid this, even though the assignment of \texttt{s} 
is inserted into the body code of \texttt{A}, we do not delete the node of 
\texttt{s}, but simply mark it as \textit{artifact}. 
Artifacts are nodes concerning size-type operations that are needed to avoid creating 
dependencies between real nodes. After each simplification, we perform 
a test to determine if any of the artifact nodes can be removed. In the example above, if it turns out that \texttt{B} is a view of \texttt{A}, in the third simplification step, the assignment of \texttt{B} is moved to the body code of \texttt{A}, after which \texttt{C} depends directly on the node of \texttt{A}, so we can remove the artifact node of \texttt{s}.
    \item Finally, the view and insert operations are simplified. 
    View nodes are merged with the node of their data owner by inserting 
    their assignment into the body code of the data owner.           
    The same idea applies to \textit{in-place} operations, 
    i.e. we insert them into the node of their \textit{data owner}. 
    However, since Rockmate wants to control the backward execution, 
    after each \textit{main} operation we \textit{detach} the tensor 
    to split the backward graph. This \textit{detach} operation must 
    be performed before creating different independent views, 
    otherwise we would have to \textit{detach} each view independently, 
    which is impossible in PyTorch. On the other hand, the \textit{detach} 
    operation must be performed after in-place operations because they 
    impact the data, even though these in-place operations can be applied 
    to views and not directly to the original tensor.
    PyTorch is aware of that, so it ensures it's impossible to have
    in-place operations over different independent views, 
    therefore we always have a valid way to run the code and \textit{detach} 
    at the right position. An attribute \textit{in-place\_code} 
    is provided to handle the \textit{detach} operation.
\end{itemize}

\textit{Artifact} nodes which survived until the end of simplification
will be considered as \textit{soft-dependencies} when 
at the moment we generate a topological order for the 
final forward-backward graph, only to ensure the owner of an artifact 
always comes 
before any of its users. Apart from that they are removed. 
Thus, we do not create explicit dependencies, 
but since ILP follows the topological order for the first computation of 
each node, the size information is computed before it is used and is not 
forgotten. 

At the end of the simplification, we end up with a 
\textit{forward} graph, where each node consists of exactly one 
\textit{real} operation that creates a \textit{data}, 
and some secondary operations on it. As for the topological order, 
we follow the original order by using the unique number we assigned 
to each variable. 
Thus, we keep the order in which \textit{jit} performed the \textit{forward} 
code of the original module. Finally, although we will not discuss it in detail, 
we also handle random operations.

\begin{figure}[hbtp]
    \caption{
        Number of nodes in the basic forward graph compared to         the simplified one.
    }
    \begin{center}
        \begin{tabular}{| c | c c |}
            \hline          &
            Basic graph     &
            Simplified graph            \\
            \hline
            GPT2(nlayers=2) & 164 & 35  \\
            \hline
            Resnet101       & 346 & 211 \\
            \hline
            Regnet\_x\_32gf & 245 & 173 \\
            \hline
            MLP Mixer       & 203 & 124 \\
            \hline
            nn.Transformer  & 161 & 51  \\
            \hline
        \end{tabular}
    \end{center}
\end{figure}

\begin{figure}[hbtp]
    \begin{center}
        \caption{
            Simplified forward graph of a GPT2 with 2 transformer blocks,
            generated by PGB.
        }
        \includegraphics[height=0.99\textheight]{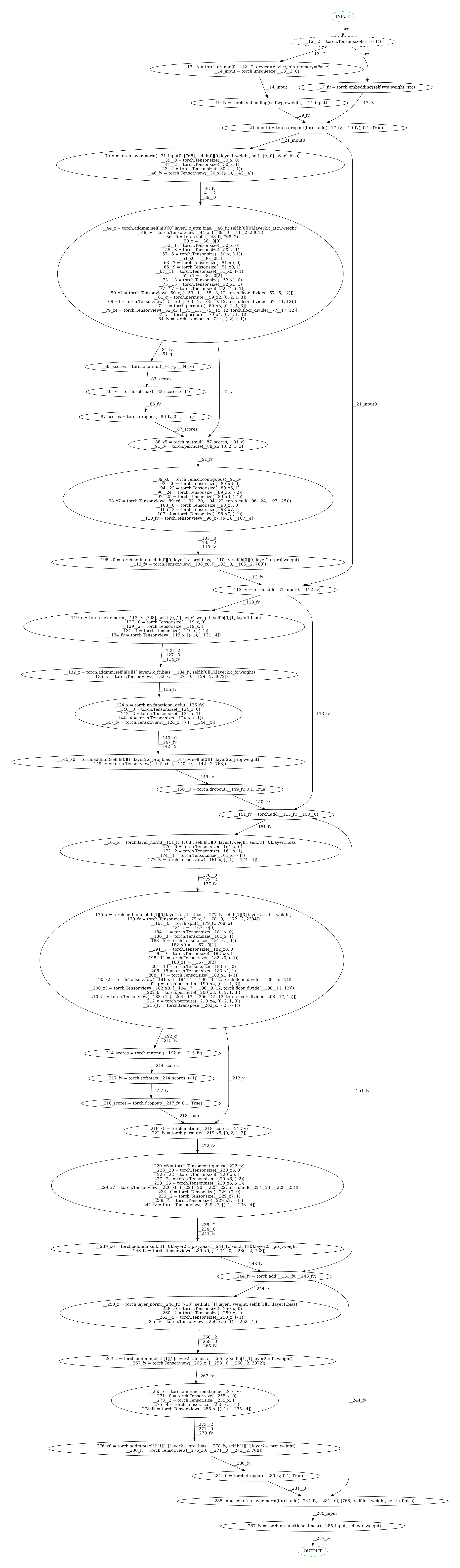}
    \end{center}
\end{figure}


\subsection{Cut the forward graph}

The subsequent steps of Rockmate rely on a list of blocks in order to
apply rk-Checkmate to each block independently. We cut the simplified
forward graph into a sequence of blocks. The cuts are made along the
1-separators of the graph: these are nodes whose removal result in a
disconnected graph. We obtain the list of separators by using a
variant of Breadth First Search (BFS). Note that although we cut the
simplified graph, a first draft of the separators list is computed
before we perform the simplifications. We mark potential separators as
\textit{protected} to avoid oversimplifying the graph, which could
break its overall structure. Since \textit{torch.add} operations are
simplified by default, without this protection, all residual edges
would traverse the entire graph from input directly to output. As a
result, the simplified graph would consist of a single large block
with many undesirable dependencies. The \textit{protected} nodes,
which are potential separators, bypass the \textit{cheap}
simplification step. At the end of the simplification step, the list
of separators is recomputed, as it may have changed due to other
simplifications.

\begin{figure}[hbtp]
    \caption{
        Cut simplified forward graph of a GPT2 with 2 transformer blocks,
        generated by PGB.
    }
    \includegraphics[width=0.85\textwidth]{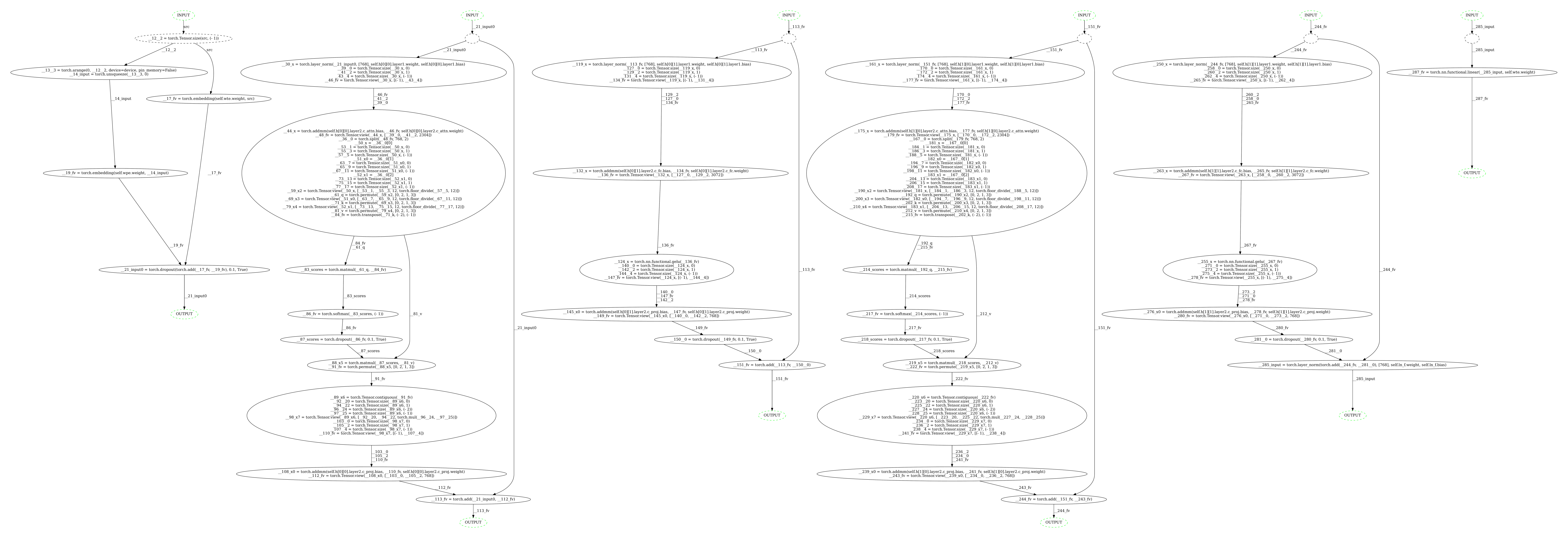}
\end{figure}

\subsection{Recognition of Similar Blocks}

To reduce solving time, we would like to avoid solving the same ILP
multiple times on identical blocks. For example, GPT2 consists of $n$
attention blocks interleaved with $n$ feedforward blocks, and the
resolution of this ILP $n$ times on identical instances should be
avoided. We developed a tool to generate classes of identical blocks.

\begin{itemize}
\item First, we provide a function to anonymize simplified graphs.
  Given a simplified graph, we build a parser that maps all target
  names to numbers starting with 1 and also anonymizes the parameter
  names.
\item To test if two blocks are identical, we compare their anonymized
  versions. This is done by visiting each node in turn, following some
  topological order (which should be the same if the blocks are
  identical). For each pair of nodes, we compare their attributes: the
  code, but also the information about each variable, including its
  type and shape.  The lists of parameters should also have the same
  anonymized names (i.e. they are used in the same nodes), but also
  the same data types and shapes.

\item After building the equivalence classes, we directly build the
  \textit{forward}+\textit{backward} graph for each unique anonymous
  graph, and translate it back to un-anonymize and
  get the \textit{forward}+\textit{backward} of each block.  In
  Rockmate, we solve the ILP once for each equivalence class, and the
  resulting \textit{R} and \textit{S} matrices can be shared across
  identical blocks.
\end{itemize}

\begin{figure}[hbtp]
    \caption{
        Total number of blocks in the cut simplified graph compared
        to the number of unique blocks.
    }
    \begin{center}
        \begin{tabular}{| c | c c |}
            \hline          &
            Chain length    &
            Unique blocks                      \\
            \hline
            GPT2            & nlayers$>1$ & 5
            \\
            \hline
            Resnet101       & 38          & 13 \\
            \hline
            Regnet\_x\_32gf & 27          & 11 \\
            \hline
            MLP Mixer       & 27          & 6  \\
            \hline
            nn.Transformer  & 3           & 3  \\
            \hline
        \end{tabular}
    \end{center}

In the GPT2 implementation used in the experiments, there are
only five unique blocks. Each transformer layer is automatically
divided into two blocks, corresponding to a \ti{attention} part and a
\ti{MLP} part. The five unique blocks are: the input
pre-processing, a typical \ti{attention} block, a typical
\ti{MLP} block, the last \ti{MLP} block which is
slightly different due to simplification, and the final output
post-processing.  Therefore, regardless of the number of transformer
blocks, the overall solving time for rk-Checkmate is constant.
\end{figure}

\begin{figure}[hbtp]
    \caption{
        An anonymized version of an attention block from GPT2
    }
    \hspace{-.5cm}
    \includegraphics[width=0.55\textwidth]{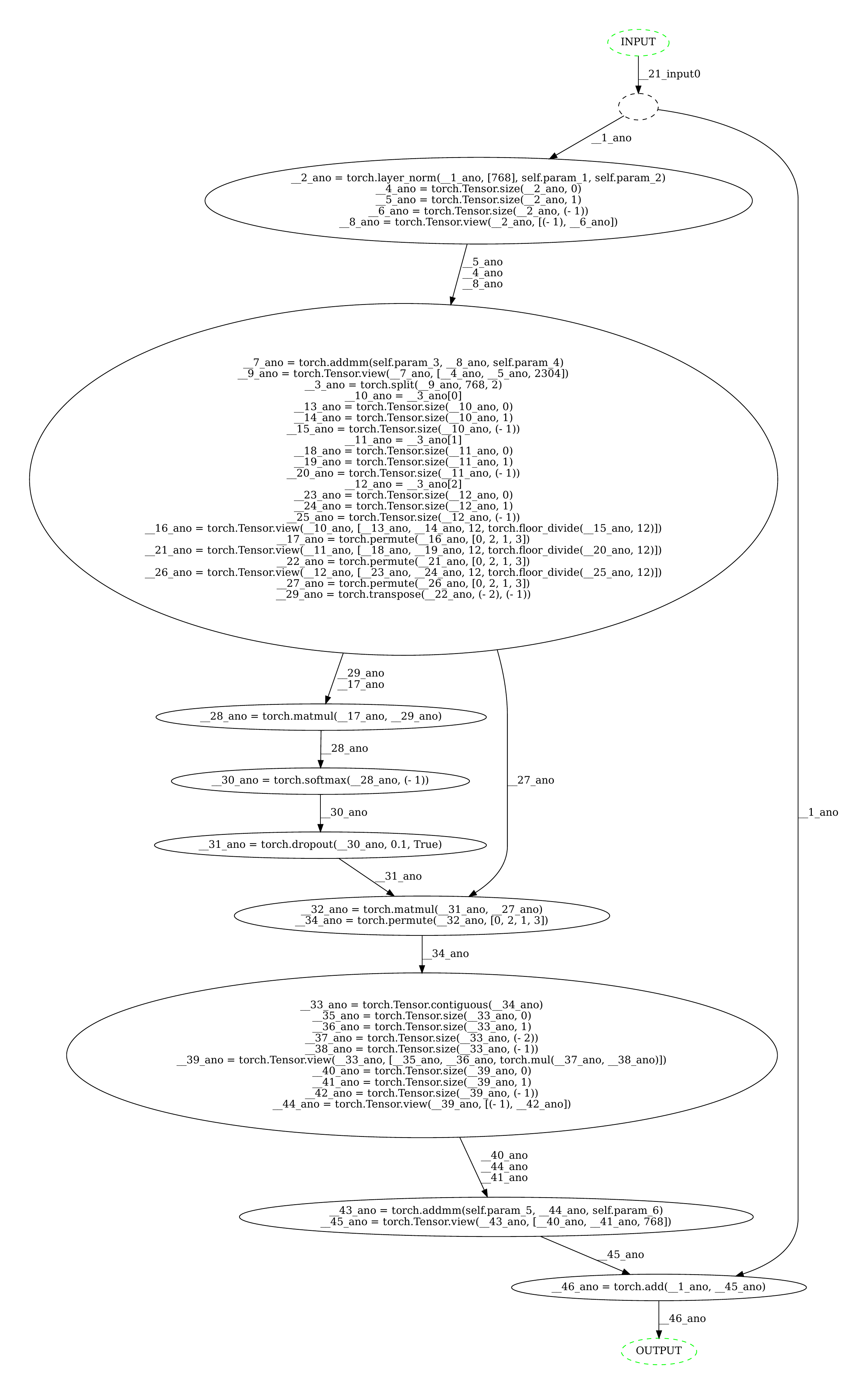}
\end{figure}

\subsection{Inspection and backward nodes}

Since we now have exactly one data defined per node, we can associate
one backward operation with each forward node, whose
code is given by \texttt{\_\textless
  target\textgreater.backward(\textless
  target\textgreater.grad)}\footnote[3]{ The \texttt{\_target} (with
  an underscore) refers to the variable before \textit{detaching},
  while \texttt{target} (without) is the \textit{detached} one.  } \\

In rk-Checkmate two categories of nodes are introduced:
\textit{Computation} nodes (\textit{C\_nodes}) and \textit{Data} nodes
(\textit{D\_nodes}). A \textit{C\_node} represents an operation that
takes a certain amount of time to execute with a certain amount of
memory overhead. It is either a \textit{Forward} or a
\textit{Backward} node. A \textit{D\_node} represents an item stored
in memory. \textit{D\_nodes} can be deleted to free memory, and
\textit{C\_nodes} can be recomputed to restore \textit{D\_nodes},
although it takes some time. . There are three types of
\textit{D\_nodes}: \textit{tensor.data} attribute,
\textit{tensor.grad} attribute, and what we call \textit{phantoms},
which are intermediate results (\textit{saved\_tensors}) stored in the
\textit{grad\_fn} attribute.

We consider each node of the simplified forward graph one at a time in
topological order. To create the \textit{C} and \textit{D\_nodes} we
need to do an inspection, \ie to run the code several times to measure
time and memory usage\footnote[4]{ For the memory usage, since we are
  assuming Rockmate is being used on GPUs, we can trust
  \textit{torch.cuda.memory\_allocated}, which is not the case on
  CPUs. rk-GB raises a warning when used on a CPU and skips the
  inspection part. }.  First we run the \textit{forward} code
(including the body part) and measure how much memory has been
allocated, then we set the \textit{.data} attribute to
\textit{torch.empty(0)} to free the memory\footnote[5]{ If it exists, we also enforce the
  \textit{.\_base.data} attribute to \textit{torch.empty(0)}, because
  it is a view of the \textit{.data} (otherwise we would not be able
  to free memory)}.

The memory successfully freed is the memory used by the
\textit{tensor.data}'s \textit{D\_node}. If some memory was not freed,
it means that the node has \textit{phantoms}. Phantoms are
intermediate values stored in \textit{tensor.grad\_fn} in preparation
for \textit{backward}.

To handle phantoms properly, we handle any \textit{.\_saved\_tensors}
or \textit{.variable} attributes. We use the
\textit{grad\_fn.next\_functions} attribute to recursively open the
\textit{grad\_fn} graph. The \textit{.variable} attribute is an
explicit reference to a variable (e.g. the input of the
\textit{forward} operation), while \textit{\_saved\_tensors} are
either views of known tensors or original tensors. The opening of
\textit{grad\_fn} is crucial for three reasons:

\begin{itemize}
    \item
First, we need to open \textit{grad\_fn} to properly build \bw
\ti{C\_nodes} dependencies.  Indeed, an input of the \textit{forward}
operation is needed to perform the \ti{backward} operation if and only
if there is a reference to it in the \ti{grad\_fn}.  Consider a first
example:
          \texttt{
              B = torch.addmm(bias,A,weight)
          } \\
          In such a case, \textit{A.data} is indeed necessary to run
          \textit{B.backward} and a reference to it can be found
          in \textit{B.grad\_fn}.
          Therefore, \textit{B}'s \textit{backward}
          \textit{C\_node} depends on \textit{A}'s \textit{data}
          \textit{D\_node}. Now consider a second example: \\
          \texttt{
              C = torch.add(A,B)
          } 

Here, given \textit{C.grad}, both \textit{A.grad} and \textit{B.grad}
can be computed without \textit{A.data} or \textit{B.data}, so that
we do not want \bw \Cn of \ti{C} to depend on \ti{A} and \ti{B}'s
\ti{data} \ti{D\_nodes}. Note that \ti{autograd} always checks
the shape of the \ti{.data} attribute of inputs, even if they are not used. To solve
this problem, we introduce two types of dependencies (\ie edges of the
graph): \ti{actual} and \ti{fake}.  \ti{fake} edges do not appear in
the ILP, we do not want to force the \ti{data} to be alive if it is
not used.  \ti{fake} edges are only used in Rockmate's final code
generator: in the example above, \ti{C}'s \bw \ti{fakely} depends on
\ti{A} and \ti{B}'s \ti{data}, so by the time we want to run \ti{C}'s
\ti{backward}, \ti{A.data} may already be forgotten. To pass the
\ti{autograd} shape equality check, we need to assign an empty tensor
with the correct shape to \ti{A.data}. To avoid wasting memory for
this, we use
          \texttt{
              A.data = torch.ones(1)\\
              \phantom{A.data = torc}
              .expand(\textless numel(A)\textgreater)\\
              \phantom{A.data = torc}
              .view(\textless shape(a)\textgreater)
          } \\

          Using this trick we allocate only 512 octets, whereas: 
          \texttt{
              A.data = torch.empty(\textless shape(a)\textgreater )
          } \\
          would allocate as much memory as the original
          \textit{A.data} and cancel our efforts to let the solver
          free \ti{A}'s data to reduce memory footprint.

\item In the previous paragraph we explained the user's viewpoint:
  given a tensor, we want to find its dependencies. Let us now take
  the input perspective. If a \ti{\_saved\_tensor} is a view of the
  input, it means that
          \texttt{
              input.data = torch.empty(0)
          }
  will not free anything. To free a \ti{data} we must put the
  \textit{data} attribute of all the tensors having the same
  \textit{data\_ptr} to \textit{torch.empty(0)}, including the
  \textit{\_saved\_tensors} which refer to it. The correct way to
  forget a \ti{data} is therefore
  \texttt{ input.data = torch.empty(0)
    ;
    \\ input\_view.data = torch.empty(0) ;
    \\ user.grad\_fn.next\_functions[0][0].\_saved\_mat1.data=   torch.empty(0) }

Similarly, after recomputing \textit{input}'s data, all
\textit{\_saved\_tensors} that refer to it are rebuilt. Even if it is
not directly the same \textit{data}, but a view of it, we take care to
rebuild it properly, including operations that affect the strides.
\textit{autograd} can store any view of the data, but there is no way
to guess if the stride will be affected.  However, we use all known
views of the data that are not phantoms, and we find one that is
compatible with the \ti{\_saved\_tensor}. Therefore, for a node, we
need the names of the phantoms that are views of it.
   
%
%

\item As mentioned before, phantoms can be either references to
  existing tensors or original tensors. If original tensors are found
  in \textit{grad\_fn}, there will be a difference between the memory
  generated during \textit{forward} and the memory freed when
  forgetting \textit{.data}. In this case we create a
  \textit{phantom} \ti{D\_node}. This node has exactly one dependency
  (to the \fw \ti{C\_node}) and one user (the \bw \ti{C\_node}).


\end{itemize}

By inspecting \textit{forward} execution and \textit{backward} when
\textit{requires\_grad} is set, we obtain the memory and time
attributes of all the mentioned nodes. Finally, there is another
special \Cn. It applies to everything related to \textit{loss}. The
special \textit{loss} \Cn depends on the model's output \ti{data} \Dn
and is required by the model's output \ti{grad} \ti{D\_node}. This
node has no \textit{code} attribute, it is a placeholder to split the
final schedule into \textit{forward} and \textit{backward} phases. In
the rk-Checkmate generated schedule, the \bw part starts as soon as
this \Cn is computed. Furthermore, in rk-Checkmate's ILP, in addition
to the generation constraint: $memory\_allocated < M_{peak}$, we can
control that at the moment when the special \textit{loss\_node} is
computed, $memory\_allocated < M_{save}$.

Let us conclude with two final remarks
\begin{itemize}
    \item In addition to \ti{actual} and \ti{fake} edges, we introduce
      \ti{global} edges. For example, the first \ti{C\_nodes} of a
      block \textit{global} depend on the \textit{data} \Dn of the
      previous block. These edges assist the final code
      generator. Using these edges, we have the entire \fw+\bw graph.
\item Measuring GPU memory usage with cuda is accurate, but running
  the same operation twice in different contexts can result in
  different amounts of memory being allocated. Therefore, the final
  execution may allocate more or less memory than we predicted. This
  is due to the way cuda allocates memory, trying to minimize
  \ti{memory fragmentation}. But it's usually very small.
\end{itemize}

\begin{figure*}
    \centering
    \caption{
        Final product of rk-GB for a GPT2 with 2 transformer blocks. \\
        \textit{Forward C\_nodes} are in blue, \bw ones in violet,
        \textit{D\_nodes} in gold, special nodes in green.
    }
    \includegraphics[width=1.03\textwidth]{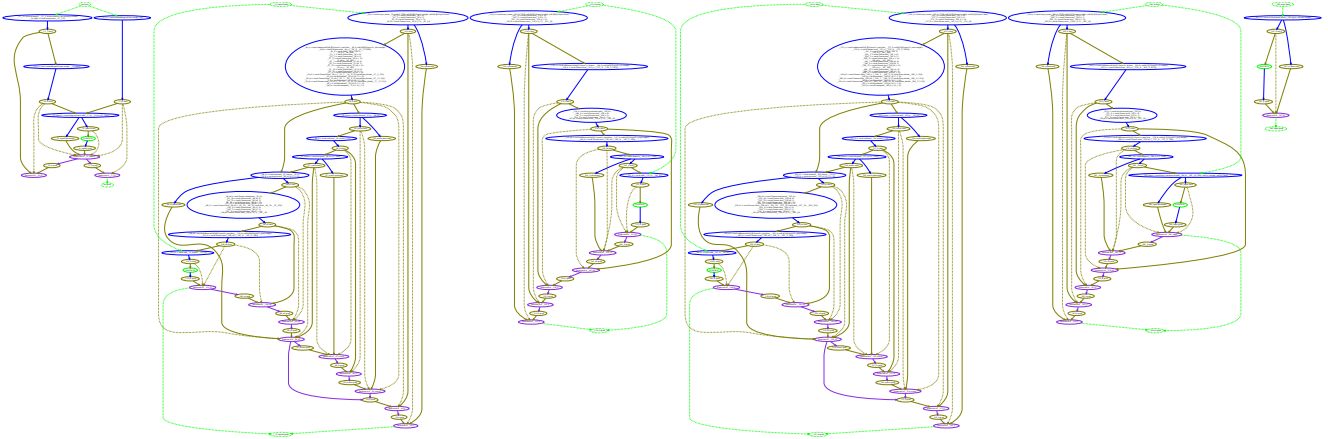}
\end{figure*}

\section{rk-Checkmate: ILP details}\label{sec:appendix:ilp}

\subsection{Graph and objective}

As it was discussed in section~\ref{sec:rk_checkmate}, at block level we find an
optimal re-materialization  strategy given memory budget via proposed rk-Checkmate
algorithm. 

Note that in Checkmate paper, they assume only one output is generated from each operation across the
forward and backward graph. In PyTorch, such an assumption is not feasible in
general cases: \texttt{tensor.grad\_fn} tends to generate the gradients of all
relevant input tensors. Therefore, our rk-Checkmate is based on a graph where each
computation node can produce multiple data nodes. Moreover, if \texttt{tensor}
is used in computation of different \texttt{target}'s, \texttt{tensor.grad}
may be generated in any backward node of \texttt{target}.

Input to the rk-Checkmate is a \texttt{CD\_graph} built by rk-GB.  
\texttt{CD\_graph} is a directed acyclic graph,
which containts:
\begin{itemize}
    \item $D$ data nodes $\{v_1,\dots,v_D\}$ and $T$ computational nodes $\{w_1,\dots,w_T\}$,
    \item edges of type $v_d \rightarrow w_t$ and $w_t \rightarrow v_d$ that show
dependencies between computational operations and data. 
For example, $v_d$ is used to perform computation $w_t$,
and computation $w_t$ outputs data $v_d$ as a result.
One computational node can have several incoming data nodes and vice versa.
\end{itemize}

To find an optimal re-materialization strategy for one block
given memory budget and computation-data dependencies described with \texttt{CD\_graph}, 
we solve an integer linear programming (ILP) problem, which minimizes
computational costs required for propagation through the block given
feasibility and memory constraints. 

Denote by $stage_{t-1\rightarrow t}$ a period,
which starts after the result of computation $w_{t-1}$ is obtained 
for the first time and ends when the computation $w_{t}$ is firstly performed.
During one stage several computations from $\{w_{t'}\}_{t'\le t}$ and deletions might happen.

{\bf The solution of ILP provides a schedule $\mathbf{R}$} (low-triangular binary matrix $T\times T$) that
determines which computations should be performed during each stage.

\begin{equation*}
    R_{t,t'}=
    \begin{cases}
      1,\quad \text{if we compute $w_{t'}$ during the $stage_{t-1\rightarrow t}$}\\
      0,\quad \text{otherwise}
    \end{cases}\,.
\end{equation*}

Each stage can be seen as a sequence of steps,
such that during one $step_{t'-1\rightarrow t'}$ one computation $w_{t'}$ is done (or not
if the schedule doesn't require that, i.e. if $R_{t, t'} = 0$) and some tensors are deleted.

{\bf Also, the solution of ILP provides an information $\mathbf{S}$} about data nodes saved during each stage. 

\begin{equation*}
    S_{t,(t', d)} =
    \begin{cases}
      1,\quad \text{if during  $stage_{t-1\rightarrow t}$ 
      an output data tensor $v_d$ of computation $w_{t'}$ is saved}\\
      0,\quad \text{otherwise}
    \end{cases}\,.
\end{equation*}

Consider all edges in \texttt{CD\_graph} that connect computation nodes with their
children data nodes, 
$${ChildrenOfComp:=\{(t', d) |\ v_d \in children(w_{t'}), t'=1,\dots,T\}},$$ 
and let their number equals $|ChildrenOfComp| = E^{t\rightarrow d}$. Then $S$ can be seen as binary matrix of size 
$T\times E^{t\rightarrow d}$.

Thus, {\bf given memory budget, ILP finds schedule $R, S$ such that computational costs} 
$$\sum_{1 \le t'<t\le T} C_{t'}R_{t,t'}$$ 
{\bf are minimized given feasibility and 
memory constraints (where $C_{t'}$ is a cost of computation $w_{t'}$).}
Now let us take a closer look to the constraints.









\subsection{Feasibility constraints}
Consider all edges in \texttt{CD\_graph} that connect data nodes with their
parent computation nodes
$$ParentsOfData:=\{(t', d) |\ w_{t'} \in parents(v_d), d=1,\dots,D\}$$ 
$$ChildrenOfData:=\{(t', d) |\ w_{t'} \in children(v_d), d=1,\dots,D\}$$ 
then a set of edges, which connects each data node with its children and parent 
computation nodes, can be expressed as
$${ChildrenParentsOfData:=ChildrenOfData\cup ParentsOfData}$$ 
and let their number equals $|ChildrenParentsOfData| = E^{t\rightarrow d\rightarrow t}$.

Let also introduce a binary matrix $P$ of size $T\times D$, where 

\begin{equation*}
    P_{t, d} =
    \begin{cases}
      1,\quad \text{if we have data tensor $v_d$ in memory after the end of $stage_{t-1\rightarrow t}$}\\
      0,\quad \text{otherwise}
    \end{cases}\,.
\end{equation*}
Note that $P_{t, d} \ge S_{t, (t', d)} / |parents(v_d)|$.

The following constraints for ILP should be hold 
\begin{itemize}
    \item $\sum\limits_{t=1}^T\sum\limits_{t'=t+1}^TR_{t,t'}=0$: ensures that we can recompute only operations that have been executed during previous stages.
    \item $\sum\limits_{e}^{}\sum\limits_{t=1}^{t''-1}S_{t,e}=0$, where $e=(t'', d), e\in ChildrenOfComp$: ensures that before the first computation, data cannot be saved.
    \item $\sum\limits_{d=1}^D\sum\limits_{t=1}^{t'}P_{t, d} = 0,$ where $t' = \min\{t''| w_{t''} \in parents(v_d)\}$: ensures data tensor $v_d$ isn't stored before the execution of first coputation that contributes to its value.
    \item $\sum\limits_{t=1}^T R_{t,t} =T$: ensures that $w_t$ is executed at the end of $stage_{t-1->t}$. 
    \item $\sum\limits_{t=1}^{T} R_{t, t_{loss}} = 1,$ where $t_{loss}$ is an index of node that computes the loss: ensures that the loss is computed only once during the forward-backward phase. 
    \item $S_{t, e} \le P_{t, d},$ where $e=(t', d)\in ChildrenOfComp$ and $t,t'=1,\dots,T$
    \item $S_{t+1, e} \le S_{t, e} + R_{t, t'},$ where $e=(t', d)\in ChildrenOfComp$ and $t=1,\dots,T-1$
    \item $R_{t, t'} \le R_{t, t''} + S_{t, e},$ where $t'\in children(d), e=(t'', d), e\in ChildrenOfComp$, and $t=1,\dots T$: ensures that all computations $w_{t'}$ which are required for generation of data $v_d$ are present, where $v_d$ is an input data for computation $w_t$. 
\end{itemize}


\subsection{Memory constraints}



Let $ChildrenOfData[d]$ denotes a set of indices of computation nodes
$w_t$, which requires tensor $d$.

And let $ParentsOfData[d]$ returns a set of indices of computation nodes
$w_t$, which generates tensor $d$.

We remind that each stage can be seen as a sequence of steps, such that during one step 
one computation (or not if the schedule doesn't require that) and some tensors are deleted.

To represent the presence of certain tensors at different steps $step_{t'-1\rightarrow t'}$ of each
stage $stage_{t-1\rightarrow t}$, we introduce the following variables:

$create_{t,d,t'}\in\{0,1\}, \forall t'\in ParentsOfData[d]$:
whether tensor $v_d$ is created during $step_{t'-1\rightarrow t'}$ at stage $stage_{t-1\rightarrow t}$.

$delete_{t,d,t'}\in\{0,1\}, \forall t'\in (ParentsOfData[d]+ChildrenOfData[d])$:
whether tensor $d$ is deleted during $step_{t'-1\rightarrow t'}$ at stage $stage_{t-1\rightarrow t}$.

Let us define expression for $t=1,\dots,T$ and $(t', d)\in ChildrenParentsOfData$: 
$$alive[t,d,t'] = P_{t,d}+\sum_{t''\le t'}create_{t,d,t''} - \sum_{t''\le t'}delete_{t,d,t''} \in\{0,1\}.$$


A tensor $v_d$ is either alive or deleted immediately after the computation of parent nodes:
$$alive[t,d,t']+delete_{t,d,t'}\geq R_{t,t'},$$

A tensor $v_d$ is retained during $stage_{t-1\rightarrow t}$ if it is alive during the last posssible step
of $stage_{t-2\rightarrow t-1}$: $$alive[t,d,t'] = P_{t,d}, t' =
    max(ParentsOfData[d], ChildrenOfData[d])$$

A tensor can only be created from the parent computation: $$create_{t,d,t'}\leq
    R_{t,t'}$$

A tensor should be deleted if it would not be needed or saved in the current
stage: $$delete_{t,d,t'} = R_{t,t'} * \prod_{d' \in att(d)}(1-P_{t+1,d'}) *
    \prod_{t'' \in children(v_d) | t''>t'}(1-R_{t,t''})$$

No tensor should be alive after the final stage: $$alive[T,d,t'] = 0,\ t' =
    max(ParentsOfData[d]+ChildrenOfData[d])$$
`'
Let $U_{t, t'}$ denotes the memory saved at the end of $step_{t'-1 \rightarrow t'}$ during $stage_{t-1 \rightarrow t}$ and
$M_d$ is the memory required to store tensor $v_d$, then
$$U_{t, 1}=\sum_{d=1}^{D}M_{d} P_{t,d} + \sum_{d=1}^D{M_{d}\ create_{t,d,1}} - \sum_{d=1}^D{M_{d}\ delete_{t,d,1}} 
     $$
$$U_{t, t'} = U_{t, t'-1}+\sum_{d=1}^D{M_{d}\ create_{t,d,t'}} - \sum_{d=1}^D{M_{d}\ delete_{t,d,t'}}$$

The peak memory at $step_{t'-1 \rightarrow t'}$ during $stage_{t-1 \rightarrow t}$ is within memory
budget: $$tmpM_{t'} R_{t, t'} + U_{t, t'}+\sum_{\forall i}{M_{d} delete_{t,d,t'}}\leq
    M_{budget}$$ where $tmpM_{t'}$ is the temporary memory overhead needed in the
computation node $w_{t'}$.


\section{rk-Rotor}
\label{sec:appendix:rk-rotor}

\subsection{Notations}

Assume our model is a sequence of $L$ blocks, numbered from $0$ to
$L-1$. For each block, we have $1+B$ budget options, where option $0$
does not save any intermediate data, and each of the other $B$ options
saves a different amount of data. We denote by $\fop{i}{o}$ the
forward computation of block $i$ with option $o$, and if $o > 0$,
$\bop{i}{o}$ is the corresponding backward computation. Since
$\fop{i}{0}$ does not store any intermediate data, we consider that it
does not have a corresponding backward computation.

The input activation of $\fop{i}{}$ is $\act{i}$, and its output is
$\act{i+1}$. Similar to the Rotor paper~\cite{rotor-RR}, for each
option $o>0$, we denote by $\bact{i}{o}$ the union of $\act{i}$ and of
all the intermediate data generated by $\fop{i}{o}$. For ease of
notation, we will also use $\act{i}$ and $\bact{i}{o}$ to denote the
size of the corresponding data.

For any computation, we use $\mtmp{\cdot}$ to denote the temporary memory
usage of this computation: this is the amount of memory that needs to be
available for this computation to succeed, and that is released
afterwards. We also use $\exec{\cdot}$ to denote the running time of an
computation.  As an example, since the input and output data need to be
in memory, the memory usage for running $\fop{i}{0}$ is $\act{i} + \act{i+1} +
    \mtmp{\fop{i}{0}}$, and this takes time $\exec{\fop{i}{0}}$.

\subsection{Formulation}

We denote by $\opt(s, t, m)$ the optimal execution time for computing
the sequence from block $s$ to block $t$, assuming that the input
$\act{s}$ will be kept in memory. There are two possible cases for the
start of this computation:
\begin{itemize}
    \item If block $s$ is only computed once in this sequence, then it is
          computed with one of the $\fop{i}{o}$ options for $o > 0$ so that it
          is possible to perform the backward computation. This requires to
          have at least $\mtmp{\fop{i}{o}}$ available memory for the
          forward and at least $\mtmp{\bop{i}{o}}$ available for the backward.
          The corresponding execution time is $\exec{\fop{i}{o}} +
              \exec{\bop{i}{o}}$, and the memory available for the rest of the
          computation is $m-\bact{i}{o}$. The best choice is given by:
          \begin{equation}
              \opt_1(s, t, m) = \min_{\text{valid option $o$}}\exec{\fop{i}{o}} +
              \exec{\bop{i}{o}} + \opt(s, t, m-\bact{i}{o}) \label{eqn:rk-rotor:fe}
          \end{equation}
          In this equation, an option is considered valid if the temporary
          memory requirements for the forward and backward computations are
          satisfied.
    \item If block $s$ is computed more than once, then its first
          computation does not need to keep any intermediate data. It is thus
          computed with $\fop{s}{0}$, and the choice now is about which is the
          next activation to be kept in memory. Let us denote by $i$ the index
          activation kept in memory, so that activations $\act{s+1},
              \act{s+2}, \dots, \act{i-1}$ are discarded just after being used. It
          is possible to compute $\act{i}$ by performing $\fop{s}{0},
              \fop{s+1}{0}, \dots, \fop{i-1}{0}$. Once this activation is computed
          and stored in memory, optimizing the rest of the computation becomes
          a subproblem: we need to compute the optimal execution time from
          block $i$ to $t$. Afterwards, since no activation was stored between
          blocks $s$ and $i$, this corresponds to another subproblem, from $s$
          to $i$. The best choice is given by:
          \begin{equation}
              \opt_2(s, t, m) = \min_{\text{valid choice $i$ with $s < i<t$}}
              \exec{\fop{s}{0}}+\exec{\fop{s+1}{0}} +\cdots+ \exec{\fop{i-1}{0}}
              + \opt(i, t, m-\act{i}) + \opt(s, i, m) \label{eqn:rk-rotor:fn}
          \end{equation}
          In this equation, a choice is considered valid if the temporary
          memory requirements for all computations $\fop{s}{0}, \fop{s+1}{0},
              \dots, \fop{i-1}{0}$ are satisfied.
\end{itemize}
In both cases, if there is no valid choice, the corresponding $\min$
value is considered to be $+\infty$. Finally, the optimal decision for
our problem is computed with:
\begin{equation}
    \opt(s, t, m) = \min \big(\opt_1(s, t, m), \opt_2(s, t, m)\big) \label{eqn:rk-rotor:all}
\end{equation}

Additionally, if $s=t+1$, only the first case can be considered, but
this time the rest of the computation is empty. We can thus compute
$\opt(s, s+1, m)$ for all $s$ and all $m$. The resulting algorithm is
close to the Rotor algorithm, using the updated
equation~\eqref{eqn:rk-rotor:fe}, and is provided in
Algorithm~\ref{alg:rk-rotor}.

\newcommand{\RETURN}{\textbf{return}}

\begin{algorithm}
\caption{rk-Rotor for $L$ blocks with memory $m$.}
\label{alg:rk-rotor}
\begin{algorithmic}[1]
\FOR {$m = 1, \dots, M$}
 \FOR {$k = 1, \dots, L$}
  \FOR {$s = 1, \dots, L+1 -d$}
      \STATE Compute $\opt(s,s+k, m)$ with equation~\eqref{eqn:rk-rotor:all}
    \ENDFOR 
  \ENDFOR
\ENDFOR
\STATE \RETURN{} $\text{rk-Rotor-Build}(\opt, 1, L+1, m-\act{0})$ \COMMENT{Alg.~\ref{alg:rk-rotor:rec}}
\end{algorithmic}
\end{algorithm}

\begin{algorithm}
  \caption{$\text{rk-Rotor-Build}(\opt, s, t, m)$ -- Computation of the schedule}
\label{alg:rk-rotor:rec}
\begin{algorithmic}
  \IF{$\opt(s, t, m) = \infty$}
  \STATE \RETURN{} Infeasible
  \ELSIF{$s = t+1$ and $\opt(s, t, m)=\opt_1(s, t, m)$ with option $o$}
  \STATE \RETURN{} $(\fop{s}{o}, \bop{s}{o})$
  \ELSIF{$\opt(s, t, m) = \opt_2(s, t, m)$ with choice $i$ (equation~\eqref{eqn:rk-rotor:fn})}
  \STATE \RETURN{} $(\fop{s}{0}, \fop{s+1}{0}, \dots, \fop{i-1}{0}, \text{rk-Rotor-Build}(\opt, i, t, m - \act{i}), \text{rk-Rotor-Build}(\opt, s, i, m))$
  \ELSE
  \STATE $o \gets$ option such that $\opt(s, t, m) = \opt_1(s, t, m)$ (equation~\eqref{eqn:rk-rotor:fe})
  \STATE \RETURN{} $(\fop{s}{o}, \text{rk-Rotor-Build}(\opt, s+1, t, m - \bact{s}{o}), \bop{s}{o})$
  \ENDIF
\end{algorithmic}
\end{algorithm}

\newpage
\section{rk-Exec}
\label{sec:appendix:rk-exec}

Rockmate's final re-materialization schedule is a list of operations, either compute or forget. rk-Exec takes care of executing this schedule properly. It creates a new \texttt{nn.Module} that produces exactly the same results (both data and gradients) while respecting the requested budget. The schedule given to rk-Exec refers to \texttt{C\_nodes} and \texttt{D\_nodes}.

\subsection{Computation}

Remember that a \texttt{C\_node} consists of a main assignment that creates the \texttt{.data}, and a \texttt{body\_code} that contains secondary statements about shapes, views, and in-place operations. By default in PyTorch, during forward execution, \texttt{autograd} puts in output's \texttt{grad\_fn} all the information needed to go back directly from the loss to input's gradients. The principle of memory saving is to control the backward and how intermediate activations are saved. To prevent \texttt{autograd} from creating the whole computational graph in output's \texttt{grad\_fn}, rk-Exec \textit{detach} each tensor after computing it, so that \texttt{grad\_fn} only keeps track of the last operation. Consider the following example\\

\texttt{
    a = torch.linear(input,...) ; \\
    b = torch.relu(a) ; \\
    c = b.view(...) ; \\
    d = torch.linear(c,...) ; \\
    d.relu(inplace=True) ; \\
    e = d.view(...)\\
    output = torch.add(c,e) ;
} \\
For the forward we have : \\
\begin{itemize}
    \item For C\_node a \\
          \texttt{
              \_a = torch.linear(input,...) ; \\
              a = \_a.detach().requires\_grad\_() ;
          }
    \item For C\_node b, viewing operations are done after \texttt{detach} \\
          \texttt{
              \_b = torch.relu(a) ; \\
              b = \_b.detach().requires\_grad\_() ; \\
              c = torch.Tensor.view(b,...)
          }
    \item For C\_node d, in-place operations are done before \texttt{detach} \\
          \texttt{
              \_d = torch.linear(c) ; \\
              \_fv1 = torch.relu\_(\_d) ; \\
              d = \_d.detach().requires\_grad\_() ; \\
              fv1 = d ; \\
              e = torch.Tensor.view(fv1,...)
          }
    \item For C\_node output\\
          \texttt{
              output = torch.add(c,e)
          }
\end{itemize}

We always name with an underscore the variable before detaching, we call it the \textit{proxy}. Based on this the backward computation is: \\
\texttt{
    \_\textless var\_name\textgreater.backward(\textless var\_name\textgreater.grad)
}

\subsection{Deletions}

Remember that there are three types of \texttt{D\_nodes}:
\begin{itemize}
    \item To free a \texttt{tensor.data}, we assign all views that refer to it to \texttt{torch.empty(0)}. In the example above, to free \textit{d's data} \texttt{D\_node} we set \texttt{d.data}, \texttt{fv1.data} and \texttt{e.data} to 0. But as mentioned in the rk-GB appendix, this also includes views that are stored in users' grad\_fn as \texttt{\_saved\_tensors}.           For example, to free \textit{b's data} \texttt{D\_node} you must perform \\
          \texttt{
              \_d.grad\_fn.next\_functions[0][0].\_saved\_mat1 = torch.empty(0)
          }
    \item To free a \texttt{phantoms} \texttt{D\_node} we just need to perform \\
          \texttt{
              del \_\textless var\_name\textgreater
          } \\
 It will delete the \texttt{grad\_fn}, but the variable defined by the detach operation (\texttt{\textless var\_name\textgreater}, without the underscore) will keep the \texttt{tensor.data} alive. So we just forget about the phantoms.     \item To release a \texttt{grad} \texttt{D\_node}, all we need is \\
          \texttt{
              \textless var\_name\textgreater.grad = None
          }
\end{itemize}

\subsection{Recomputation}

To recompute a \texttt{C\_node} rk-Exec reassign the proxy, but we do not detach again. Since the variable post-detach could be mentioned directly in its user grad\_fn. We simply reassign the \textit{data} attribute. In the example above, to recompute \texttt{C\_node} we do: \\
\texttt{
    \_d = torch.linear(c) ; \\
    \_fv1 = torch.relu\_(\_d) ; \\
    d.data = \_d.data ; \\
    fv1.data = d.data ; \\
    e.data = torch.Tensor.view(fv1,...)
} \\
Furthermore, since re-materialization is the opposite of forgetting, we need to reassign the data attribute of all views, including the \texttt{\_saved\_tensors} that refer to them. Recomputing the \texttt{phantoms} and \texttt{grad} is trivial.

\paragraph*{Last issues}

\begin{itemize}
\item In the graph produced by rk-GB we introduced the notion of \textit{real} and \textit{fake} dependencies, they imply several tricks in rk-Exec. See the last part of the rk-GB appendix for explanations.

 \item To generate the code correctly, we compile the list of operations described in the schedule. That is, we generate the code one by one, but we keep track of which tensors are alive to avoid any if statements if the final code.           For example, when we need to reassign all the views of a tensor in users grad\_fn, we know which users are alive.

 \item Rockmate manipulates code that are either string or Python AST object we can execute. Therefore, rk-Exec generate a list of strings to execute.           Then we can generate a big string code for the forward, and one for the backward, finally we use Python \textit{compile} function to execute these functions without wasting time on parsing the string.

 \item We take care of random operations, in particular to be able to recompute a random function deterministically, we store random states on the first computation and restore them when needed.

   \item Note that sometimes, due to the float approximation precision, the results obtained from the original module and the new one may be a little different. But they are as close as running the original module twice. With \texttt{torch.float64} precision there are always strictly equal on the model we tested.\end{itemize}

\newpage

\section{Detailed experiments on GPT2}

\begin{center}
  \begin{tabular}{lllrrrr}
    \toprule
    Model & Input size & Algorithm &  Budget (GiB) &  Peak mem &  Makespan &  Makespan \\
    &  &  &  &  (GiB) &  mean (ms) &  std (ms) \\
    \midrule
    GPT2-large &   (2, 512) &   PyTorch &     $M_{GPU}$ &           6.708 &             457.173 &              1.207 \\
    GPT2-large &   (2, 512) &     Rotor &         0.850 &           0.846 &             593.507 &              1.789 \\
    GPT2-large &   (2, 512) &  Rockmate &         0.850 &           0.816 &             564.041 &              1.378 \\
    GPT2-large &   (2, 512) &     Rotor &         2.800 &           2.360 &             522.163 &              0.889 \\
    GPT2-large &   (2, 512) &  Rockmate &         2.800 &           2.741 &             481.650 &              0.992 \\
    GPT2-large &   (2, 512) &     Rotor &         7.600 &           6.515 &             456.971 &              1.255 \\
    GPT2-large &   (2, 512) &  Rockmate &         7.600 &           6.516 &             465.376 &              0.985 \\
    \hline
    GPT2-large &   (4, 256) &   PyTorch &     $M_{GPU}$ &           5.158 &             432.646 &              1.207 \\
    GPT2-large &   (4, 256) &     Rotor &         0.850 &           0.808 &             561.066 &              1.791 \\
    GPT2-large &   (4, 256) &  Rockmate &         0.850 &           0.795 &             531.255 &              0.732 \\
    GPT2-large &   (4, 256) &     Rotor &         1.800 &           1.473 &             526.807 &              1.402 \\
    GPT2-large &   (4, 256) &  Rockmate &         1.800 &           1.745 &             486.976 &              1.305 \\
    GPT2-large &   (4, 256) &     Rotor &         5.600 &           4.690 &             439.797 &              1.451 \\
    GPT2-large &   (4, 256) &  Rockmate &         5.600 &           4.967 &             440.410 &              1.469 \\
    \hline
    GPT2-medium &  (2, 1024) &   PyTorch &     $M_{GPU}$ &          10.782 &             480.330 &              0.975 \\
    GPT2-medium &  (2, 1024) &     Rotor &         1.000 &           1.188 &             751.478 &              1.070 \\
    GPT2-medium &  (2, 1024) &  Rockmate &         1.000 &           0.994 &             619.055 &              1.012 \\
    GPT2-medium &  (2, 1024) &     Rotor &         4.000 &           3.351 &             558.858 &              1.208 \\
    GPT2-medium &  (2, 1024) &  Rockmate &         4.000 &           3.941 &             516.758 &              0.899 \\
    GPT2-medium &  (2, 1024) &     Rotor &        11.600 &          10.326 &             494.043 &              1.097 \\
    GPT2-medium &  (2, 1024) &  Rockmate &        11.600 &          10.582 &             490.490 &              0.924 \\
    \hline
    GPT2-medium &   (4, 512) &   PyTorch &     $M_{GPU}$ &           7.407 &             430.596 &              1.206 \\
    GPT2-medium &   (4, 512) &     Rotor &         1.000 &           1.188 &             658.736 &              1.105 \\
    GPT2-medium &   (4, 512) &  Rockmate &         1.000 &           0.986 &             547.343 &              0.873 \\
    GPT2-medium &   (4, 512) &     Rotor &         2.000 &           1.824 &             532.637 &              1.686 \\
    GPT2-medium &   (4, 512) &  Rockmate &         2.000 &           1.988 &             497.298 &              0.833 \\
    GPT2-medium &   (4, 512) &     Rotor &         7.600 &           6.664 &             445.434 &              0.710 \\
    GPT2-medium &   (4, 512) &  Rockmate &         7.600 &           7.207 &             440.769 &              1.410 \\
    \bottomrule
  \end{tabular}
\end{center}
\end{document}